\documentclass{article}

\usepackage{microtype}
\usepackage{graphicx}
\usepackage{subcaption}
\usepackage{booktabs} 

\usepackage{hyperref}
\usepackage{listings}

\usepackage[accepted]{icml2026}

\usepackage{amsmath}
\usepackage{amssymb}
\usepackage{mathtools}
\usepackage{amsthm}

\usepackage[capitalize,noabbrev]{cleveref}

\theoremstyle{plain}

\theoremstyle{definition}

\theoremstyle{remark}

\usepackage[textsize=tiny]{todonotes}

\icmltitlerunning{Closing the Expression Gap in LLM Instructions via Socratic Questioning}

\begin{document}

\lstset{
  breaklines=true,        
  breakatwhitespace=false,
  basicstyle=\ttfamily\small,
  frame=single,
  columns=fullflexible,
  keepspaces=true,
  showstringspaces=false
}

\twocolumn[
  \icmltitle{Closing the Expression Gap in LLM Instructions via Socratic Questioning}



\icmlsetsymbol{equal}{*}

\begin{icmlauthorlist}
  \icmlauthor{Jianwen Sun}{nankai,sii,equal}
  \icmlauthor{Yukang Feng}{nankai,sii,equal}
  \icmlauthor{Yifan Chang}{sii,equal}
  \icmlauthor{Chuanhao Li}{shlab} 
  \icmlauthor{Zizhen Li}{nankai,sii} \\
  \icmlauthor{}{} \\
  \icmlauthor{Jiaxin Ai}{whu,sii}
  \icmlauthor{Fanrui Zhang}{sii}
  \icmlauthor{Yu Dai}{nankai}
  \icmlauthor{Kaipeng Zhang}{sii,shanda}$^{\dagger}$
\end{icmlauthorlist}

\icmlaffiliation{nankai}{Nankai University}
\icmlaffiliation{sii}{Shanghai Innovation Institute}
\icmlaffiliation{whu}{Wuhan University}
\icmlaffiliation{shlab}{Shanghai AI Laboratory}
\icmlaffiliation{shanda}{Shanda AI Research}

\icmlcorrespondingauthor{Jianwen Sun}{sunjianwen@mail.nankai.edu.cn}
\icmlcorrespondingauthor{Kaipeng Zhang}{kaipeng.zhang@shanda.com}

\icmlkeywords{Machine Learning, ICML}

\vskip 0.3in
]



\printAffiliationsAndNotice{}  

\begin{abstract}

A fundamental bottleneck in human-AI collaboration is the ``intention expression gap", the difficulty for humans to effectively convey complex, high-dimensional thoughts to AI.
This challenge often traps users in inefficient trial-and-error loops and is exacerbated by the diverse expertise levels of users.
We reframe this problem from passive instruction following to a Socratic collaboration paradigm, proposing an agent that actively probes for information to resolve its uncertainty about user intent.
we name the proposed agent Nous, trained to acquire proficiency in this inquiry policy.
The core mechanism of Nous is a training framework grounded in the first principles of information theory.
Within this framework, we define the information gain from dialogue as an intrinsic reward signal, which is fundamentally equivalent to the reduction of Shannon entropy over a structured task space.
This reward design enables us to avoid reliance on costly human preference annotations or external reward models.
To validate our framework, we develop an automated simulation pipeline to generate a large-scale, preference-based dataset for the challenging task of scientific diagram generation.
Comprehensive experiments, including ablations, subjective and objective evaluations, and tests across user expertise levels, demonstrate the effectiveness of our proposed framework. Nous achieves leading efficiency and output quality, while remaining robust to varying user expertise.
In conclusion, our research provides a systematic methodology and a new perspective for addressing the issue of ambiguous intentions in complex human-machine collaboration.

\end{abstract}

\begin{figure}[t]
    \centering
    \includegraphics[width=0.92\linewidth]{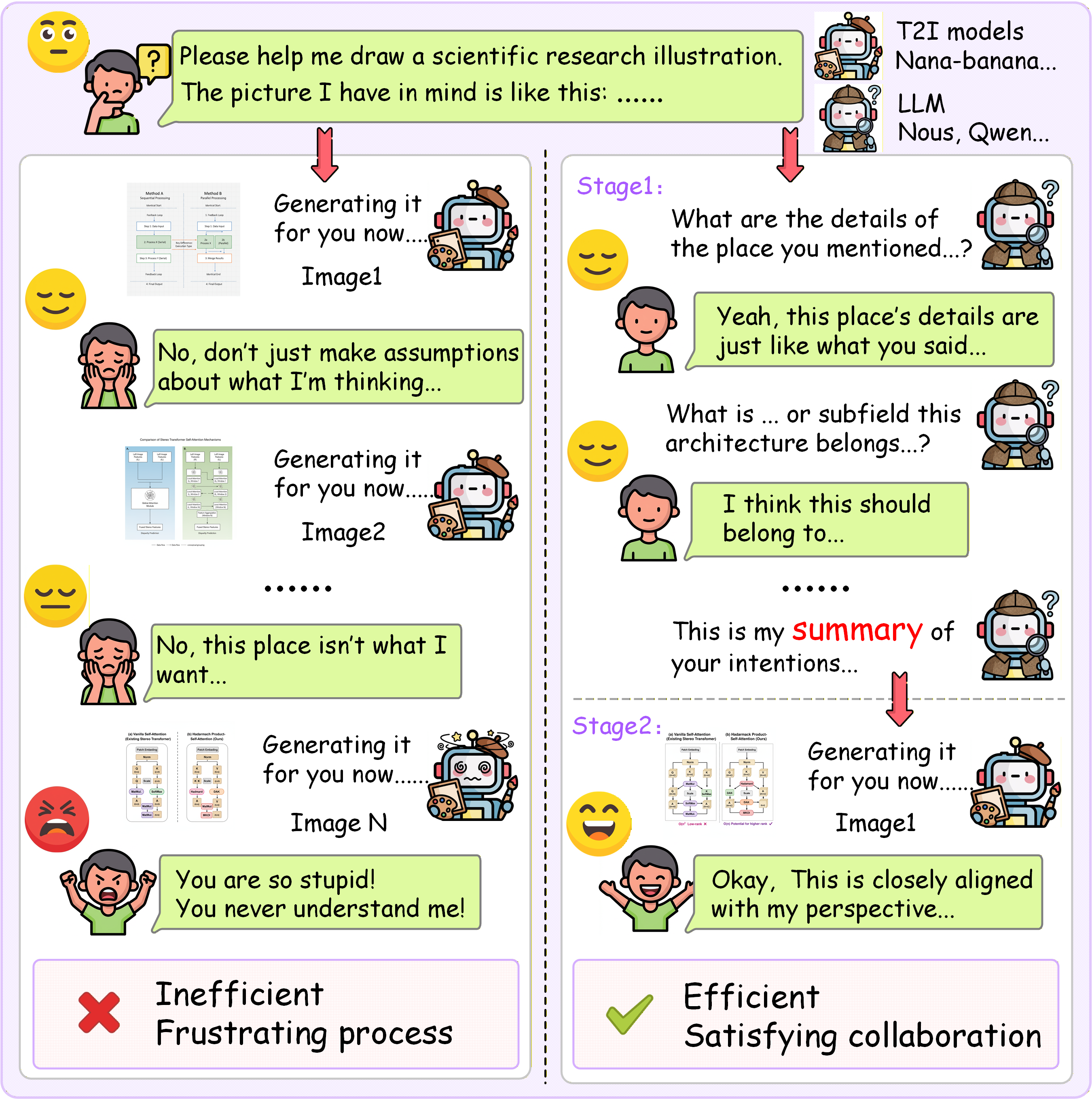}
    \caption{ Comparison of different paradigms. Left part: Generate round by round; Right part: Clarify doubts prior to generation.}
    \vspace{-2mm}
    \label{fig:teaser}
\end{figure}

\section{Introduction}
The transition of AI from an efficient tool to a true collaborative partner hinges on solving a core challenge: achieving a shared understanding with the user~\citep{liang2025samepage}.
While Large Language Models (LLMs) demonstrate remarkable fluency in text generation, their passive, instruction-following nature falters when faced with the inherent incompleteness of human intent expression~\citep{shneiderman2020human}.
This limitation is especially evident in creative and technical domains~\citep{amershi2019guidelines,ahn2023can}.
In such settings, users may hold highly innovative ideas yet struggle to articulate them with precision~\citep{ahn2022intentgap}.
When attempting to realize these ideas with AI, they often fall into a frustrating “guessing game,” which in turn forces task goals to emerge gradually and be refined through collaborative processes~\citep{amershi2014power}.
The gap between a user’s high-dimensional mental model and their ability to convey it in a machine-readable format has been described as the ``intention gap,''~\citep{Vanessa2024} which forces collaboration into inefficient trial-and-error loops~\citep{buccinca2020proxy}.
As a result, the entire burden of precise articulation falls on the human, and this paradigm is fundamentally unsustainable for complex tasks.

Our research stems from a core insight: Why must humans always painstakingly teach the AI, instead of the AI intelligently guiding the human? We advocate for a paradigm shift: envisioning AI not as a passive follower, but as an agent actively bridging this gap~\citep{zhou2020designing,lee2022co}. Inspired by the Socratic method, we treat it not merely as pedagogy but as a model for collaborative discovery~\citep{kim2023socratic}, The collaborative approach shown in Figure ~\ref{fig:teaser}. A Socratic agent does not simply await commands; it formulates strategic questions to systematically resolve its uncertainty about the user’s goal~\citep{krishna2022visual,roberts2020measurement}. Each question-answer turn becomes a deliberate act of information seeking, designed to maximize convergence toward a shared, high-fidelity understanding~\citep{holstein2020participatory,activate_1,activate_2,activate_3}.

To this end, we introduce Nous, an agent designed to acquire proficiency in an optimal inquiry policy. The central mechanism of Nous is a training framework grounded in the first principles of information theory~\citep{cover2006elements,entropy_1,entropy_2}. Within this framework, we define the information gain from dialogue as an intrinsic reward, formally equivalent to the reduction of Shannon entropy over possible task specifications. By relying on this objective and computationally tractable signal, Nous avoids dependence on costly human preference annotations or external reward models~\citep{christiano2018supervising,lee2022survey,lambert2022illustrating}.

To validate this framework, we select scientific diagram generation as our testbed, a prototypical instance of the intention gap. The task is both high-dimensional and logically structured, providing objective criteria for evaluation while remaining sufficiently challenging~\citep{green2020using,liu2023chartllm}. Building on this, we construct an automated simulation pipeline to generate a large-scale, preference-based dataset tailored to this setting~\citep{rafailov2023grpo}. Finally, we conducted comprehensive experiments and evaluations, which demonstrated the effectiveness of our method. Moreover, we attempted to expand its application scenarios, and experiments in collaborative creation suggest the potential of the generalization performance of our method~\citep{lee2022co,zhou2023survey}.

\vspace{-0.7mm}
Our contributions are summarized as follows:

\vspace{-0.7mm}
(1) \textbf{Nous}, An intelligent agent that instantiates the Socratic interaction paradigm has obvious advantages in both interaction efficiency and generation quality. 

\vspace{-0.7mm}
(2) \textbf{An information-theoretic framework}, using dialogue-driven information gain as an intrinsic reward, eliminating the need for human annotation or external reward models. 

\vspace{-0.6mm}
(3) \textbf{An automated large-scale simulation pipeline}, generating dialogue strategy learning data to support scalable training and evaluation.

\begin{figure*}[t]
    \centering
    \includegraphics[width=\linewidth]{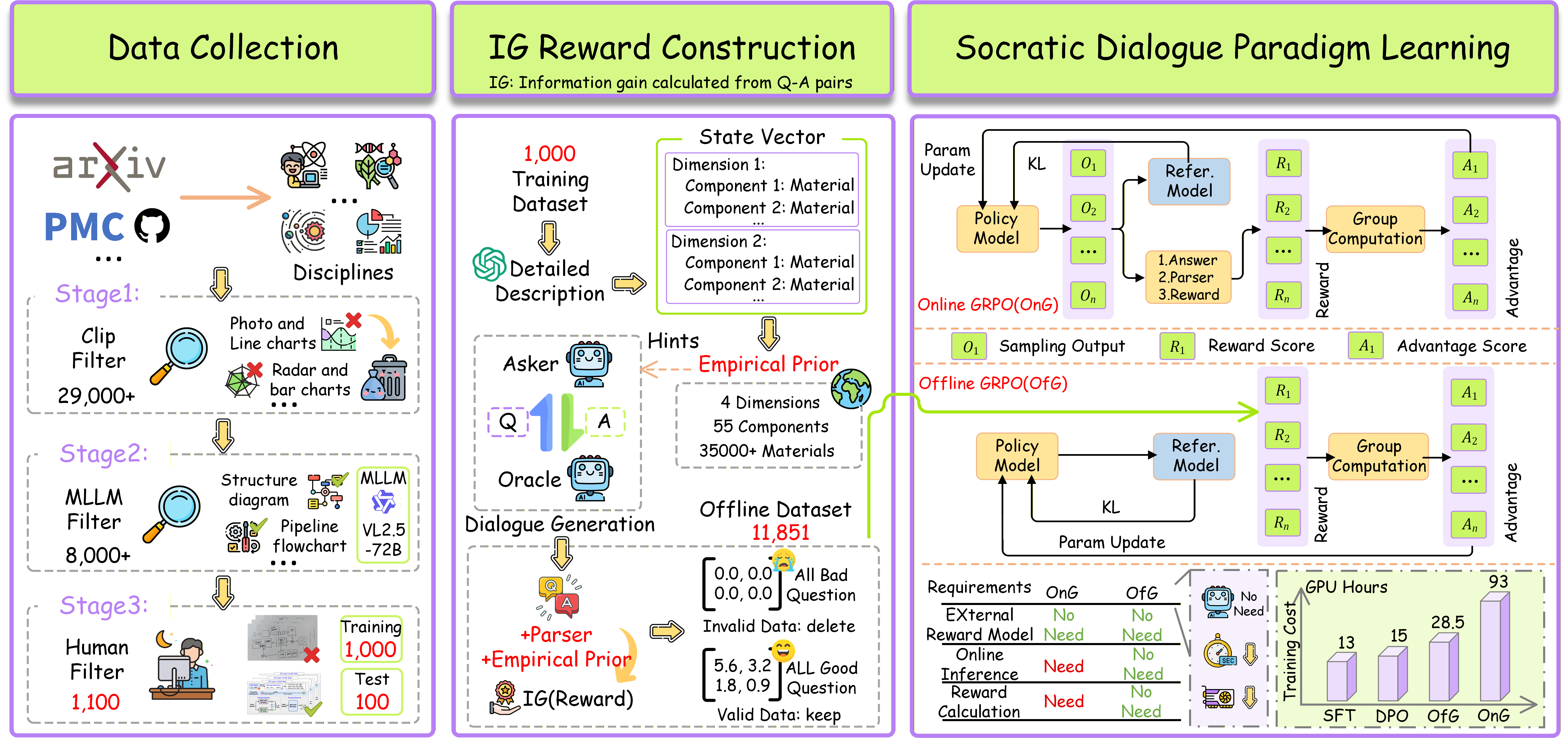}
    \caption{
        \textbf{The multi-stage curation pipeline for the dataset and the details of model training.} 
        We began with a raw dataset of approximately 1 million figures downloaded from scientific papers in different fields on arXiv and PMC. This dataset was first filtered using the CLIP model to remove data plots (such as bar charts and line graphs), resulting in 29,000 images. Next, we used the Qwen-2.5-VL-72B model to retain true schematic diagrams, reducing the dataset to 8,000 images. Finally, three PhD students conducted a manual review to ensure the relevance, clarity, and quality of each figure, resulting in a final dataset of 1,100 images. From this curated dataset, 1,000 figures were used to build the empirical prior and train simulations, while 100 figures were set aside for testing. Detailed explanations regarding data distribution and open-source licenses are provided in Appendix~\ref{sec:data_details}.
    }
    \vspace{-2mm}
    \label{fig:data_pipeline}
\end{figure*}

\section{Related Work}

Our work is situated at the intersection of three key areas in AI and human-computer interaction: goal-oriented dialogue, active learning, and large language model alignment.

\textbf{Goal-Oriented Dialogue Systems.} Traditional task-oriented dialogue (TOD) systems, typified by datasets like MultiWOZ~\citep{woz1,woz2,woz3,woz4}, excel in explicit slot-filling tasks such as booking flights~\citep{young2013pomdp, wen2017network}. However, these systems operate on a convergent retrieval'' paradigm, assuming a fixed set of slots to retrieve a pre-existing database entry. In contrast, creative design tasks involve divergent construction,'' where the goal is to create a novel specification from scratch, requiring dynamic attribute combinations rather than static forms(Details in Appendix~\ref{sec:Related Work}). While recent LLM-based approaches explore proactive clarification in QA~\citep{cq-gen,huang2025codeclarify} or future-planning~\citep{xu2024futureclarification}, most remain passive recipients of instructions. Our work moves beyond both traditional TOD and passive LLMs: Nous navigates a combinatorially complex specification space to resolve ambiguity, transforming the agent into an active inquirer for open-ended construction.

\textbf{Active Learning and Optimal Experiment Design.}  
Uncertainty reduction by asking questions is rooted in active learning and optimal experiment design~\citep{settles2009active, lewis1994sequential}.  
Prior dialogue-policy research has incorporated entropy reduction as a signal for clarification~\citep{zhang2021dialogpolicy,su_1}, and recent studies formalize question quality directly via expected information gain~\citep{pan2024informativequestions, geishauser2021what, xing2024}.  
However, these methods typically target static datasets or constrained “20-questions” benchmarks.  
Our contribution is to extend this principle to dynamic dialogue for creative design: instead of selecting a data point, Nous learns to generate natural language questions that probe a latent goal space.  
Training this generative policy with entropy reduction as a real-time reward bridges classical theory with modern LLM interaction~\citep{active_pref_inference2023, seed_grpo2025, dynamic_profile_modeling2025}.

\textbf{LLM Alignment and Preference-Based Learning.} 
Aligning LLMs with human intent is a central challenge. 
Preference-based methods such as RLHF~\citep{christiano2017deep, ouyang2022training}, PPO-based optimization~\citep{schulman2017ppo}, and more recent approaches like GRPO~\citep{rafailov2023grpo}, DPO~\citep{rafailov2023dpo}, and RLAIF~\citep{bai2022constitutional, lee2024rlaif} rely on costly preference labels or heuristic feedback. 
Our method offers another perspective solution: we define an intrinsic reward from information gain, bypassing external reward models and the associated annotation cost. By applying offline RL~\citep{levine2020offline, kostrikov2022offline} on automatically generated preferences, Nous avoids proxy misspecification while maintaining principled grounding in task structure, offering a complementary path for alignment in structured co-creative tasks.


\section{Methodology}
\label{sec:framework}
Our methodology is presented in three parts. First, we establish a formal information-theoretic framework, deriving an intrinsic and tractable reward signal from first principles (Sec.~\ref{sec:info_framework}). 
Next, we detail the complete offline training pipeline, which includes an automated simulation for preference data generation and the offline policy optimization algorithm. (Sec.~\ref{sec:policy_optimization}). 
Finally, we introduce the baseline models used for our comparative experiments (Sec.~\ref{sec:baselines}).

\vspace{-1mm}
\subsection{Information-Theoretic Framework}
\label{sec:info_framework}
To learn an effective inquiry strategy, the agent requires a quantitative metric for guidance. Drawing from classical information theory, we define a reward signal based on information gain, which measures the informational value of each question-answer turn. We validate our method on the scientific chart generation task, where the dialogue is modeled as a process of reducing epistemic uncertainty over a structured state space. The information gain from a user's response is formally defined as the Kullback-Leibler (KL) divergence between the posterior and prior belief states(the agent's probability distribution over user intentions). We prove this metric simplifies to the reduction in the system's Shannon entropy. This provides an intrinsic reward signal, directly calculable from the agent's belief state, for optimizing the inquiry policy without requiring a separate, pre-trained reward model.

\vspace{-1mm}
\paragraph{Formalizing the Diagram Specification Space.}
We begin by defining the object of our inquiry. A complete scientific diagram specification, denoted by $\mathcal{G}$, is conceptualized as a point in a high-dimensional, discrete state space. A diagram specification is represented by a set of $N$ attributes, $\mathcal{V} = \{V_1, V_2, \dots, V_N\}$. Each attribute $V_i$ takes a value $v_i$ from its finite, discrete domain $\mathcal{S}_i$. A complete and valid diagram specification is an instantiation $\mathbf{g} = (v_1, v_2, \dots, v_N)$ where $v_i \in \mathcal{S}_i$ for all $i \in \{1, \dots, N\}$. The attributes are designed to be comprehensive, covering aspects such as overall layout ($V_{\text{layout}}$), color palettes ($V_{\text{color}}$), the number and types of components ($V_{\text{num\_comp}}, V_{\text{comp\_type}}^{(k)}$), and interconnections ($V_{\text{conn}}^{(i,j)}$).

\vspace{-1mm}
\paragraph{Quantifying and Decomposing Epistemic Uncertainty.}
At any turn $t$ in the dialogue, the agent's knowledge about the user's desired diagram is captured by a probabilistic belief state, $P_t(\mathcal{G})$. For computational tractability, we assume the attributes $V_i$ are conditionally independent given the dialogue history $\mathcal{H}_t$(A detailed analysis of the necessity and impact of this design can be found in the Appendix~\ref{sec:limitation}). While this is a simplifying assumption, we argue it is a tractable and effective first-order approximation, as the greatest reduction in uncertainty, particularly in early dialogue, comes from resolving major, orthogonal attributes (\textit{e.g.}, overall layout, number of components)~\citep{d_3}.

This allows the joint distribution to be factorized:
\begin{equation}
    P_t(\mathcal{G}) = P(V_1, \dots, V_N \mid \mathcal{H}_t) = \prod_{i=1}^{N} P(V_i \mid \mathcal{H}_t).
    \label{eq:factorize}
\end{equation}
The agent's initial belief state, $P_0(\mathcal{G})$, is an empirical prior derived from a large-scale corpus $\mathcal{D}$ of existing diagrams, where each prior probability is estimated via maximum likelihood:
\begin{equation}
    P_0(V_i = v_j) = \frac{\mathrm{Count}_{\mathcal{D}}(V_i = v_j)}{|\mathcal{D}|}.
    \label{eq:prior}
\end{equation}
The total uncertainty of the system is the Shannon entropy of the belief state $P_t(\mathcal{G})$. A critical consequence of the independence assumption is that the total entropy decomposes into a sum of marginal entropies:
\begin{equation}
    \mathcal{H}\!\left(P_t(\mathcal{G})\right)
    \!=\! -\! \sum_{\mathbf{g} \in \mathcal{G}} \! P_t(\mathbf{g}) \log_2 \! P_t(\mathbf{g})
    \!=\! \sum_{i=1}^{N} \mathcal{H}\!\left(P_t(V_i)\right),
    \label{eq:total_entropy}
\end{equation}
where $\mathcal{H}\!\left(P_t(V_i)\right) = -\sum_{v_j \in \mathcal{S}_i} P_t(V_i = v_j)\,\log_2 P_t(V_i = v_j)$. This decomposition is crucial, as it allows us to track uncertainty on a per-attribute basis.

\paragraph{Belief State Update and Reward Function.}
The dialogue proceeds as a sequence of belief state updates. An answer $A_t$ is mapped by a semantic parser $f$ to evidence $\mathcal{E}_t$, which imposes hard constraints on a subset of attributes $\mathcal{V}_{\mathcal{E}_t}$. In our simulation, $f$ is implemented as a few-shot prompted LLM, whose parsing accuracy is ensured by the Oracle's templated responses, providing a reliable signal for reward calculation. This updates the belief from a prior $P_t$ to a posterior $P_{t+1}$ via Bayesian conditioning. For any constrained attribute, the posterior becomes a deterministic Kronecker delta function, $P_{t+1}(V_i = v_j) = \delta_{jk}$, while unconstrained attributes remain unchanged.

We define our reward signal $r_t$ as the \emph{reduction in Shannon entropy} of the belief state:
\begin{equation}
    r_t \equiv IG(A_t) \;=\; \mathcal{H}\!\left(P_t(\mathcal{G})\right) - \mathcal{H}\!\left(P_{t+1}(\mathcal{G})\right).
    \label{eq:kl_divergence}
\end{equation}
Intuitively, this quantity measures the informational value of the user’s answer. From an information-theoretic perspective, the expected value of this entropy reduction equals the mutual information between $A_t$ and $\mathcal{G}$, which can be written as an expectation over a KL divergence:
\begin{flalign}
\mathbb{E}[IG(A_t)] \!=\! I(A_t;\mathcal{G}) \!=\! \mathbb{E}_{A_t}\!\left[ D_{KL}\big(P_{t+1}(\mathcal{G}) \| P_t(\mathcal{G})\big) \right].
\hspace{-14pt}
&& \label{eq:mutual_info}
\end{flalign}
Thus maximizing information gain is identical to maximizing the reduction of uncertainty.

By substituting the entropy decomposition from Eq.~\ref{eq:total_entropy} into Eq.~\ref{eq:kl_divergence}, we derive a tractable reward function:
\begin{align}
    r_t
    &= \sum_{i=1}^{N} \left( \mathcal{H}\!\left(P_t(V_i)\right) - \mathcal{H}\!\left(P_{t+1}(V_i)\right) \right).
    \label{eq:derivation_step}
\end{align}
Under our hard-constraint update model, the posterior entropy $\mathcal{H}\!\left(P_{t+1}(V_i)\right)$ becomes zero for any newly constrained attribute $V_i \in \mathcal{V}_{\mathcal{E}_t}$, and remains unchanged for all other attributes. Therefore, the sum in Eq.~\ref{eq:derivation_step} simplifies to include only the terms for the resolved attributes:
\begin{equation}
    r_t \;=\; \sum_{V_i \in \mathcal{V}_{\mathcal{E}_t}} \mathcal{H}\!\left(P_t(V_i)\right).
    \label{eq:final_reward}
\end{equation}
This final equation states that the utility of an answer is the sum of the prior entropies of the attributes it clarifies. This signal is intrinsic, computationally efficient, and provides a robust foundation for optimizing the agent's inquiry policy. It is worth noting that we employ an unweighted sum of entropy reduction. We avoid manual weighting because Shannon entropy naturally embeds an ``implicit statistical weighting'': attributes with higher variance in the empirical prior yield larger information gain, automatically guiding the agent to prioritize statistically significant uncertainties without subjective heuristics.

\subsection{Offline Policy Optimization}
\label{sec:policy_optimization}

With a defined reward signal, we can now train the agent's inquiry policy. Our approach is a fully offline process, which enhances stability and computational efficiency. It consists of two main stages: first, we generate a large-scale, static dataset of preference-ranked inquiries through simulation; second, we use this dataset to train the policy via an offline reinforcement learning algorithm.

\paragraph{Automated Preference Data Generation}
\label{sec:datagen}
Our training process relies on a large-scale preference dataset, $\mathcal{D}_{\text{pref}}$, which we generate through an automated simulation framework. This simulation requires two key components: a ``empirical prior" to provide prior probabilities (as in Eq.~\ref{eq:prior}) and a set of ground-truth tasks. We construct both from a high-quality corpus of scientific diagrams, curated through a multi-stage filtering pipeline detailed in Figure~\ref{fig:data_pipeline}.

Within the simulation, an ``Oracle" agent, holding a ground-truth specification from our curated set, provides answers to inquiries proposed by multiple candidate models. The information gain for each inquiry is calculated via Eq.~\ref{eq:final_reward}, yielding a training sample $\{p, \{r_1, \dots, r_k\}, \{R_1, \dots, R_k\}\}$, consisting of a prompt, a group of candidate responses, and their corresponding reward scores.

\paragraph{Offline Adaptation of  Group Relative Policy Optimization.}
To optimize our policy $\pi_{\theta}$ on the static dataset $\mathcal{D}_{\text{pref}}$, we adapt the objective function from Group Relative Policy Optimization (GRPO) for an offline setting. While GRPO was originally proposed as an online algorithm that iteratively samples from the policy, we find its core objective is well-suited for offline training in our context.
The rationale for this offline adaptation is twofold. First, the task of ``asking a good question" is a capability already inherent in pretrained LLMs. The distribution of our generated candidate responses is therefore not expected to be drastically different from what the policy would generate, making on-policy sampling less critical. Second, using a static dataset eliminates the computational overhead of online generation, leading to a much more efficient and stable training pipeline.

For each group of responses, we first normalize the rewards into advantage estimates $A(r_i, p)$ via z-scoring within the group. This stabilizes the learning process across different prompts. Our offline algorithm then maximizes the following PPO-style clipped surrogate objective:
\begin{equation}
\begin{split}
L_{\text{Policy}}(\theta) = \mathbb{E}_{(p, r_i, A_i) \sim \mathcal{D}_{\text{pref}}} \Bigl[ \min \bigl( \rho_i(\theta) A_i, \quad \quad \\
\text{clip}(\rho_i(\theta), 1-\epsilon, 1+\epsilon) A_i \bigr) \Bigr],
\end{split}
\label{eq:grpo_loss}
\end{equation}
where the ratio $\rho_i(\theta) = {\pi_{\theta}(r_i|p)}/{\pi_{ref}(r_i|p)}$ measures the policy change against a reference policy $\pi_{ref}$. The clipping function $\text{clip}(\cdot)$ limits this ratio to a trusted region, preventing overly aggressive and destabilizing policy updates.

To further regularize the policy and ensure it does not deviate excessively from the pre-trained base model, we incorporate a KL-divergence penalty, leading to the final loss function:
\begin{equation}
L_{\text{total}}(\theta) = L_{\text{Policy}}(\theta) - \beta D_{KL}(\pi_{\theta}(\cdot|p) \,||\, \pi_{ref}(\cdot|p)),
\label{eq:total_loss}
\end{equation}
where $\beta$ is a hyperparameter controlling the strength of the KL penalty. The log-probabilities $\log \pi(r|p)$ are computed autoregressively. To ensure the policy is only trained on its generation, we apply a loss mask so that the gradients are backpropagated only through the tokens of the response $r$, not the prompt $p$.

\subsection{Contrasting Methods for Ablation Study}
\label{sec:baselines}


To rigorously evaluate the effectiveness of the \textbf{offline GRPO (OfG)} paradigm, we will use several other key baselines to train Nous for comparison in the experiments.

\textbf{Supervised Fine-Tuning (SFT)}: A baseline model fine-tuned only on the highest-reward (prompt, response) pairs from our dataset. This helps isolate the contribution of preference-based optimization over simple imitation learning.
\textbf{Direct Preference Optimization (DPO)}: To compare against a prominent pairwise preference learning method, we implement a DPO baseline. DPO optimizes the policy to directly increase the log-probability ratio of preferred to dispreferred responses, using only the best and worst responses from each group.
\textbf{Online GRPO (OnG)}: To validate the efficiency and stability of the offline approach, we also train a model using an online GRPO pipeline. This involves an initial SFT warm-up, followed by an iterative process of sampling responses from the policy, calculating their rewards, and updating the policy. All training methods ultimately include an SFT to train their ability for final integrated description.

\section{Experiments}


We conduct a comprehensive set of experiments to evaluate our proposed framework. Our evaluation is designed to answer four key questions:
(1) Does our information-theoretic approach lead to more efficient interactions compared to established baselines?
(2) Does higher interaction efficiency translate to superior quality in the final generated artifact?
(3) Is the information gain-based reward signal the critical component of our framework's success?
(4) How robust is the learned inquiry policy to variations in user expertise?

\begin{table*}[t]
\centering
\small
\caption{Experimental results of interaction efficiency, training resource consumption, and dynamic information gain.}
\label{tab:efficiency_cost_revised}
\setlength{\tabcolsep}{8pt} 
\begin{tabular}{lcccccccc}
\toprule
& Turns & Total IG & Resource & \multicolumn{5}{c}{Information Gain Dynamics at Turn (↑)}  \\
\cmidrule(lr){5-9}
Model & (↓) & (↑) & hours(↓) & Turn 1 & Turn 5 & Turn 10 & Turn 15 & Turn 20  \\
\midrule
Nous (OfG) &20.3 & \textbf{120.5} & 28.5 &10.4 & 66.6 & \textbf{99.1} & \textbf{113.7} & \textbf{120.5} \\
Nous (OnG) & 22.0 & 115.8 & 93  &7.8  & 59.4 & 88.4 & 107.2 & 114.3 \\
Nous (DPO) & 21.5 & 111.3 & 15  &\textbf{13.9} & 65.8 & 90.7 & 101.5 & 110.9 \\
Nous (SFT) & \textbf{17.1} & 94.3 & \textbf{13}   &12.6 & \textbf{78.1} & 90.5 & 94    & 94.3  \\
\midrule
GPT-few & 22.6 & 93.5 & N/A   &9.1  & 60.4 & 77.4 & 88.1 & 92.1 \\
GPT-zero & 26.5 & 84.8 & N/A  &11.3 & 43.2 & 59.7 & 72.7 & 78.3 \\
Qwen-few & 19.5 & 90.5 & N/A  &10.6 & 61.1 & 76.4 & 85.9 & 90.5 \\
Qwen-zero & 25.3 & 81.5 & N/A &6.6  & 48.1 & 64.4 & 77.2 & 80.3 \\
\bottomrule
\end{tabular}
\vspace{-2mm}
\end{table*}

\begin{figure*}[t]
    \centering
    \includegraphics[width=0.83\linewidth]{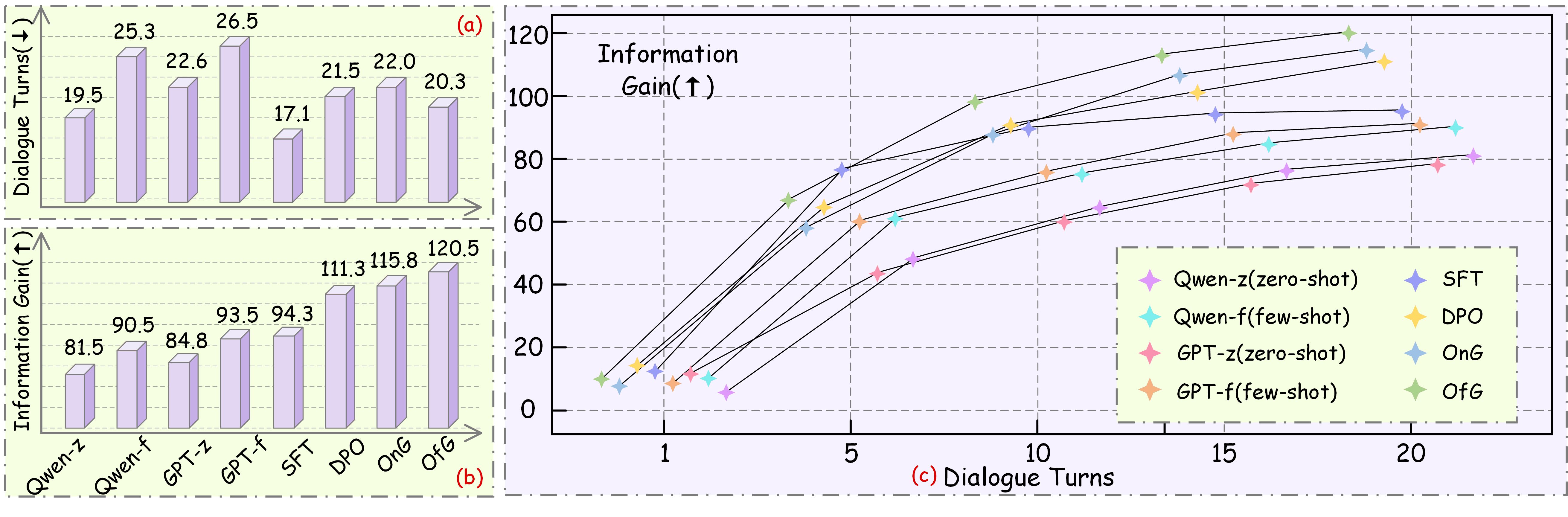}
    \caption{Experimental results of Interaction Efficiency. (a) The average number of dialogue turns for each model to complete information collection; (b) The average information gain obtained during the dialogue for each model; (c) The dynamic change of information gain during the dialogue}
    \label{fig:exp1_res}
    \vspace{-4mm}
\end{figure*}

\subsection{Experimental Setup}
\paragraph{Models Under Evaluation.}

Our primary model, Nous, is built upon Qwen3-8B and trained with full-parameter fine-tuning. 
For evaluation, we consider two groups of baselines. 
Trained Baselines: three Nous variants trained with alternative methods (SFT, DPO, OnG; see Section~\ref{sec:baselines}). 
Prompt-Based Baselines: a proprietary model (GPT-5: GPT-few ,GPT-zero) and a large open-source model (Qwen3-235B: Qwen-few, Qwen-zero), each tested under zero-shot and few-shot prompting. All prompts are instantiated using the \emph{Socratic prompting} paradigm ~\citep{socratic}, which encourages the model to ask clarifying questions before producing a figure. 
We include these as the most relevant horizontal comparison. It is worth noting that due to the novelty of the task, there are no other models or systems capable of performing such tasks except the method in~\citep{socratic}. However, we still provide the most extensive horizontal comparison experiments in Appendix~\ref{sec:horizontal}. The prompt templates and hyperparameters are also in Appendix~\ref{sec:hayper}.

\vspace{-2mm}
\paragraph{Evaluation Task and Data.}

We take the task of scientific diagram generation in human-AI collaboration as our test scenario. The test data comes from a hold-out set of 100 complex real-world diagrams (see Figure\ref{fig:data_pipeline}, for detailed sources see Appendix ~\ref{sec:vispainter}). 
For each diagram, we simulate an interaction where the agent must elicit the complete specification from an Oracle. The Oracle, which holds the ground-truth specification for a target diagram and is configured identically to the one used for generating our training data. Each dialogue begins with a generic initial request, 
``I want to create a scientific diagram," and concludes when the agent indicates it has gathered sufficient information by outputting a final, consolidated description of the diagram. This automated simulation ensures a fair, controlled, and reproducible comparison across all models.

\vspace{-4mm}
\paragraph{Evaluation Metrics.} 
We employ a multifaceted evaluation strategy to assess both the process and the outcome:
\textbf{Interaction Efficiency}: (1)We measure this by the average number of turns an agent takes to complete the dialogue, (2)and the cumulative information gain achieved throughout the interaction. Higher efficiency is indicated by fewer turns and a steeper information gain curve.
\textbf{Output Quality}: We assess the quality of the final specification from two complementary angles: (1) subjective preference scores, where the final generated diagrams are evaluated by human and AI judges through pairwise comparisons, and (2) a suite of objective, specification-based metrics that quantitatively score the generated diagrams against the ground truth.

\subsection{Main Results}

\subsubsection{Interaction Efficiency}


\textbf{Dialogue Turns and Resource Cost}: 
Table \ref{tab:efficiency_cost_revised} details the average number of dialogue turns and the associated training costs. First, all trained models complete the task in fewer turns than their non-trained counterparts. Then the performance of the SFT-trained agent shows the highest dialogue efficiency and the lowest training resources, but this brevity corresponds to the lowest total information gain among all trained models, indicating a premature and incomplete inquiry process. In contrast, the agent trained with OfG maintains a competitive turn count while requiring resources only marginally higher than DPO and significantly lower than OnG. This result highlights the scalability and cost-effectiveness of our offline training framework.

\begin{table*}[!t]
\centering
\small
\caption{Model win rate results under different tie-handling protocols: (1) ``Win": ties not counted; (2) ``W/T(0.5)": ties contribute 0.5; (3) ``W/T": ties count as 1. 
All win-rate proportions are based on 400 pairwise judgments per model pair (100 prompts $\times$ 2 judges $\times$ 2 renderers); 
the standard error of a proportion is at most 0.025, so all 95\% confidence intervals are within $\pm$0.05.}

\label{tab:subjective_win_rates}
\setlength{\tabcolsep}{6pt} 
\begin{tabular}{lcccccccccccc}
\toprule
& \multicolumn{6}{c}{4o-image-1} & \multicolumn{6}{c}{nano-banana} \\
\cmidrule(lr){2-7} \cmidrule(lr){8-13}
 Model & \multicolumn{3}{c}{Human Judge(↑)} & \multicolumn{3}{c}{GPT-5 Judge(↑)} & \multicolumn{3}{c}{Human Judge(↑)} & \multicolumn{3}{c}{GPT-5 Judge(↑)} \\
\cmidrule(lr){2-4} \cmidrule(lr){5-7} \cmidrule(lr){8-10} \cmidrule(lr){11-13}
 & \scriptsize \textbf{Win} & \scriptsize \textbf{W/T(0.5)} & \scriptsize \textbf{W/T} & \scriptsize \textbf{Win} & \scriptsize \textbf{W/T(0.5)} & \scriptsize \textbf{W/T} & \scriptsize \textbf{Win} & \scriptsize \textbf{W/T(0.5)} & \scriptsize \textbf{W/T} & \scriptsize \textbf{Win} & \scriptsize \textbf{W/T(0.5)} & \scriptsize \textbf{W/T} \\
\midrule
Nous (OfG) & \textbf{0.68} & 0.71 & 0.73 &\textbf{ 0.69 }& \textbf{0.72} & \textbf{0.76} & \textbf{0.61} & \textbf{0.66} & \textbf{0.72} & \textbf{0.55} & 0.61 & 0.66 \\
Nous (OnG) & 0.69 & \textbf{0.72} & \textbf{0.75} & 0.63 & 0.67 & 0.71 & 0.60 & 0.65 & 0.70 & 0.54 & \textbf{0.61} & \textbf{0.67} \\
Nous (DPO) & 0.59 & 0.61 & 0.64 & 0.56 & 0.61 & 0.67 & 0.57 & 0.64 & 0.71 & 0.54 & 0.59 & 0.65 \\
Nous (SFT) & 0.49 & 0.51 & 0.53 & 0.42 & 0.48 & 0.55 & 0.38 & 0.48 & 0.57 & 0.37 & 0.48 & 0.59 \\
\midrule
GPT-few & 0.45 & 0.47 & 0.50 & 0.45 & 0.52 & 0.58 & 0.34 & 0.44 & 0.54 & 0.39 & 0.47 & 0.56 \\
GPT-zero & 0.29 & 0.32 & 0.35 & 0.27 & 0.32 & 0.37 & 0.29 & 0.37 & 0.46 & 0.36 & 0.47 & 0.57 \\
Qwen-few & 0.36 & 0.40 & 0.43 & 0.35 & 0.42 & 0.48 & 0.35 & 0.42 & 0.49 & 0.32 & 0.41 & 0.49 \\
Qwen-zero & 0.23 & 0.27 & 0.30 & 0.24 & 0.28 & 0.32 & 0.26 & 0.33 & 0.40 & 0.28 & 0.36 & 0.44 \\
\bottomrule
\end{tabular}
\end{table*}

\textbf{Information Gain (IG) Dynamics}: Figure \ref{fig:exp1_res} plots the cumulative information gain against the number of dialogue turns, offering a more granular view of the inquiry strategies. The agents trained via OnG and OfG exhibit the most sustained information gain curves, demonstrating a robust ability to consistently pose high-value questions throughout the interaction. The SFT-trained agent, however, reveals a critical weakness: despite a strong start, its performance mirrors that of the non-trained models after the initial turns. They all fall into an ``information bottleneck," where the ability to ask meaningful, probing questions sharply diminishes, causing their gain curves to flatten. This empirically validates the ``frustrating guessing game" that motivated our work and underscores the necessity of a structured, goal-oriented training paradigm to overcome this limitation.

\subsubsection{Output Quality}
\label{sec:exp2}

\textbf{Subjective comparison}
We used two text-to-image backbone models (4o-image-1 and nano-banana) to generate two sets of test image data based on the final natural language summaries output by each test model. These images were evaluated twice for their drawing quality through pairwise comparisons by human reviewers and GPT-5 with reference to the test set, resulting in a total of 11,200 comparisons. Table \ref{tab:subjective_win_rates} reports the statistical results under the evaluation protocols of three tie-handling methods. It can be seen that the results from the two judging protocols were highly consistent, lending reliability to the evaluation setup. Among the trained models, Nous trained with OfG and OnG achieved the highest win rates, outperforming DPO and SFT. The non-trained baselines lagged behind, with GPT-based models generally stronger than Qwen-based ones, demonstrating the effectiveness of our method. Detailed comparison result heatmaps and case studies are provided in the Appendix~\ref{sec:data_example}.

\vspace{2mm}
\textbf{Objective metrics:}
To complement these subjective judgments with reproducible quantitative scores, we employed the VisPainter framework,a tool that converts text specifications into editable vector graphics, with examples and detailed  of its metrics provided in the Appendix~\ref{sec:vispainter} and~\ref{sec:vispainter_figure}. TWe evaluated across six dimensions: Precision, Recall, Design Error Rate, Blank Ratio, Readability, and Alignment. Weighted score is calculated by applying weights of [0.2, 0.2, 0.2, 0.05, 0.25, 0.1] to six metrics, shown in Table~\ref{tab:objective_metrics} and Figure~\ref{fig:radar}(a), highlight clear differences:
OnG and OfG perform better in terms of drawing precision, recall, and readability. This is attributed to more detailed and information-rich image descriptions. The same applies to the blank ratio; thanks to more abundant component information, more efficient space utilization is achieved. Unexpected results were observed in terms of design error rate and alignment. This is because the number of output elements is proportional to the chance of making mistakes during the drawing process, so SFT and prompt-based baseline models show higher scores in error rate and alignment. These patterns further confirm that models trained with principled inquiry signals have advantages over untrained models.

\subsection{Ablation Studies}

\paragraph{Reward Function:}

To validate the critical role of our proposed information-theoretic reward signal, we conducted ablations. We replaced it with a heuristic-based ``slot-counting" reward, which simply counts the number of specified attributes in each turn and treats all attributes equally. Using this new reward, we trained a model variant named Nous-Counting with the same OfG method on a dataset of scale, generated using the process from Section \ref{sec:datagen}.

\begin{figure*}[t]
    \centering
    \includegraphics[width=0.71\linewidth]{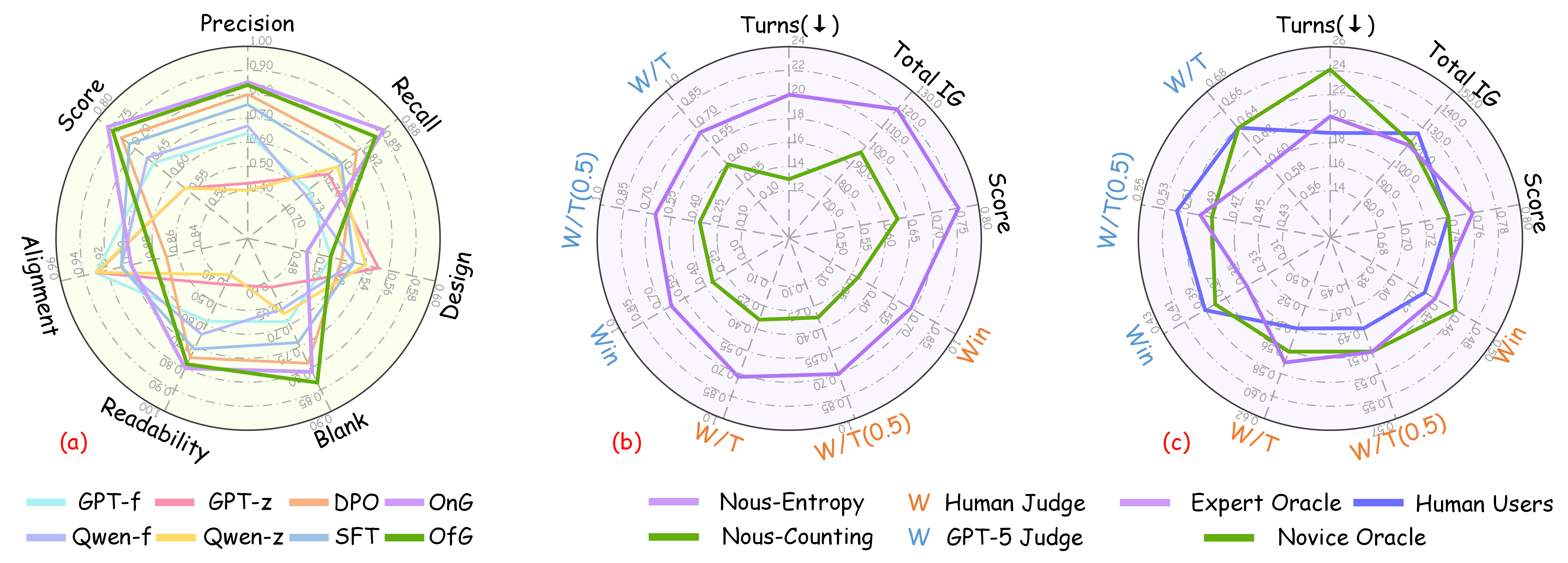}
    \caption{Visualization of experimental results. (a) Evaluation results of each model; (b) Results of ablation experiment 1; (c) Results of ablation experiment 2.}
    \label{fig:radar}
\end{figure*}

We evaluated this model under the identical experimental setup, with the results presented in Table \ref{tab:ablation_reward} and \ref{fig:radar}(b). Nous-Counting completes dialogues in fewer turns, but it achieves substantially lower information gain and final output quality. This is because the slot-counting reward encourages a greedy policy that maximizes the quantity of resolved attributes, not their informational value. The model learns to ask broad, low-impact questions rather than strategically targeting high-entropy attributes first. This study confirms that our information-theoretic reward is essential for guiding the agent to learn an inquiry strategy that is not just superficially fast, but deeply effective.

\begin{table*}[t]
\centering
\small
\caption{Results of the final generated charts using the VisPainter framework. Higher scores in each item are better, and the design error rate has also been inverted to follow the same principle.}
\label{tab:objective_metrics} 
\setlength{\tabcolsep}{7pt} 
\begin{tabular}{lccccccc}
\toprule
Model & Precision & Recall & Design & Blank & Readability & Alignment & Score \\
\midrule
Nous (OfG) & 0.83 & 0.84 & 0.51 & \textbf{0.83} & 0.79 & 0.88 & 0.76 \\
Nous (OnG) & \textbf{0.84} & \textbf{0.86} & 0.49 & 0.81 & \textbf{0.80} & 0.90 & \textbf{0.77} \\
Nous (DPO) & 0.80 & 0.81 & 0.52 & 0.78 & 0.75 & 0.87 & 0.74 \\
Nous (SFT) & 0.76 & 0.79 & 0.53 & 0.74 & 0.71 & 0.91 & 0.72 \\
\midrule
GPT-few  & 0.63 & 0.74 & 0.51 & 0.69 & 0.59 & 0.93 & 0.65 \\
GPT-zero & 0.42 & 0.77 & \textbf{0.55} & 0.61 & 0.41 & 0.93 & 0.57 \\
Qwen-few & 0.67 & 0.73 & 0.53 & 0.66 & 0.64 & 0.91 & 0.67 \\
Qwen-zero & 0.40 & 0.78 & 0.54 & 0.67 & 0.38 & \textbf{0.93} & 0.57 \\
\bottomrule
\end{tabular}
\vspace{-2mm}
\end{table*}

\begin{table*}[!ht]
\centering
\small
\caption{Quantitative evaluation of reward mechanisms and user adaptability. The table reports interaction turns, total Information Gain (IG), and win rates against human/GPT-5 judges. The upper section validates the superiority of our Information Entropy reward over a Counting baseline. The lower section demonstrates performance stability across different user levels, including simulated oracles and real humans starting from scratch (\textit{Human User\_Zero}) or partial drafts (\textit{Human User\_Draft}).}
\label{tab:ablation_reward}
\setlength{\tabcolsep}{8pt} 
\begin{tabular}{lccccccccc}
\toprule
& Turns & Total IG & Score & \multicolumn{3}{c}{Human Judge(↑)} & \multicolumn{3}{c}{GPT-5 Judge(↑)} \\
\cmidrule(lr){5-7} \cmidrule(lr){8-10}
Method & (↓) & (↑) & (↑) &\footnotesize Win & \footnotesize W/T(0.5) & \footnotesize W/T &\footnotesize Win & \footnotesize W/T(0.5) & \footnotesize W/T  \\
\midrule
Nous-Entropy  & 20.3 & \textbf{120.53} & \textbf{0.76} & \textbf{0.68} & \textbf{0.70} & \textbf{0.72} & \textbf{0.60} & \textbf{0.63} & \textbf{0.66} \\
Nous-Counting & \textbf{13.6} & 97.11  & 0.63 & 0.28 & 0.30 & 0.32 & 0.34 & 0.37 & 0.40 \\
\midrule
Expert Oracle & 20.3 & 120.53 & \textbf{0.76} & 0.41 & 0.50 & 0.59 & 0.33 & 0.49 & \textbf{0.65} \\
Novice Oracle & 24.1 & 122.47 & 0.74 & \textbf{0.42} & 0.49 & 0.57 & \textbf{0.38} & 0.51 & 0.64 \\

Human User\_Zero & 17.3 & \textbf{126.44} & 0.75 & 0.40 & \textbf{0.51} & \textbf{0.61} & 0.38 & 0.49 & 0.61 \\
Human User\_Draft & \textbf{7.6} & 42.31  & 0.74 & 0.42 & 0.50 & 0.59 & 0.37 & \textbf{0.51} & 0.64 \\

\bottomrule
\end{tabular}
\vspace{-4mm}
\end{table*}

\paragraph{User Testing:} \label{sec:User Expertise} 
To explore the performance of our system under non-ideal conditions, we  simulated roles of different levels and 10 real human users for testing. The simulated roles include: Expert Oracle, who use precise technical terms; Novice Oracle, who use vague descriptive language; and human users are recruited by us (specific information can be found in the Appendix~\ref{sec:human}). To rigorously evaluate the actual utility, we conducted human evaluations under two different settings: (1) \textit{Zero-start}, where users describe their intentions from scratch; (2) \textit{Draft-start}, where users provide an initial draft prompt containing partial information.
As shown in Table \ref{tab:ablation_reward}, Nous demonstrates strong adaptability across all user types. In the \textit{Zero-start} setting, human users required fewer turns (17.3) than the Simulator (20.3) due to the higher information density of natural language, where users tend to disclose multiple attributes per turn(Statistical data and analysis from different testers are provided in the Appendix~\ref{sec:human}). Most notably, the \textit{Draft-start} setting demonstrated a significant efficiency leap, reducing the interaction to just 7.6 turns while maintaining high generation quality (Score 0.74). This highlights a central advantage of our Socratic framework: rather than relying on flawless user input, it strategically poses follow-up questions to progressively converge on the user’s intent, proving effectively robust whether starting from vague descriptions or partial drafts.

\vspace{-2mm}
\section{Conclusion}
This paper addresses a bottleneck in human-AI collaboration: ``intention expression gap." We shift the paradigm from passive instruction-following to active, Socratic collaboration, introducing Nous, an agent that learns to resolve uncertainty about user intent through thoughtful inquiry. Our contribution is a training framework grounded in information theory, defining information gain as an intrinsic reward to eliminate costly human annotation and external reward models. We further show that Offline GRPO provides an efficient and stable path for training such agents. Experiments demonstrate that Nous achieves leading efficiency and output quality, while ablations confirm that the information-theoretic reward, rather than simple heuristics, is the decisive factor, and the agent remains robust across diverse levels of user expertise. In sum, this work presents a new methodology and perspective for resolving intent ambiguity, shifting the communication burden away from humans and moving us closer to AI partners capable of genuine collaborative thought.

\bibliography{example_paper}

@misc{liang2025samepage,
      title={On the Same Page: Dimensions of Perceived Shared Understanding in Human-AI Interaction}, 
      author={Qingyu Liang and Jaime Banks},
      year={2025},
      eprint={2505.20068},
      archivePrefix={arXiv},
      primaryClass={cs.HC},
      url={https://arxiv.org/abs/2505.20068}, 
}

@misc{Vanessa2024,
      title={Bayesian Intention for Enhanced Human Robot Collaboration}, 
      author={Vanessa and Xiaotong Zhang and Kamal Youcef-Toumi},
      year={2024},
      eprint={2410.00302},
      archivePrefix={arXiv},
      primaryClass={cs.RO},
      url={https://arxiv.org/abs/2410.00302}, 
}

@book{shneiderman2020human,
    author = {Shneiderman, Ben},
    title = {Human-Centered AI},
    publisher = {Oxford University Press},
    year = {2022},
    month = {01},
    isbn = {9780192845290},
    doi = {10.1093/oso/9780192845290.001.0001},
    url = {https://doi.org/10.1093/oso/9780192845290.001.0001},
    eprint = {https://academic.oup.com/book/41126/book-pdf/50987951/9780192659996_web.pdf},
}

@inproceedings{amershi2019guidelines,
author = {Amershi, Saleema and Weld, Dan and Vorvoreanu, Mihaela and Fourney, Adam and Nushi, Besmira and Collisson, Penny and Suh, Jina and Iqbal, Shamsi and Bennett, Paul N. and Inkpen, Kori and Teevan, Jaime and Kikin-Gil, Ruth and Horvitz, Eric},
title = {Guidelines for Human-AI Interaction},
year = {2019},
isbn = {9781450359702},
publisher = {Association for Computing Machinery},
address = {New York, NY, USA},
url = {https://doi.org/10.1145/3290605.3300233},
doi = {10.1145/3290605.3300233},
booktitle = {Proceedings of the 2019 CHI Conference on Human Factors in Computing Systems},
pages = {1–13},
numpages = {13},
location = {Glasgow, Scotland Uk},
series = {CHI '19}
}

@misc{ahn2023can,
      title={Can Large Language Models Serve as Rational Players in Game Theory? A Systematic Analysis}, 
      author={Caoyun Fan and Jindou Chen and Yaohui Jin and Hao He},
      year={2023},
      eprint={2312.05488},
      archivePrefix={arXiv},
      primaryClass={cs.AI},
      url={https://arxiv.org/abs/2312.05488}, 
}

@misc{ahn2022intentgap,
      title={Bridging the Gap Between LLMs and Human Intentions: Progresses and Challenges in Instruction Understanding, Intention Reasoning, and Reliable Generation}, 
      author={Zongyu Chang and Feihong Lu and Ziqin Zhu and Qian Li and Cheng Ji and Zhuo Chen and Hao Peng and Yang Liu and Ruifeng Xu and Yangqiu Song and Shangguang Wang and Jianxin Li},
      year={2025},
      eprint={2502.09101},
      archivePrefix={arXiv},
      primaryClass={cs.HC},
      url={https://arxiv.org/abs/2502.09101}, 
}

@article{amershi2014power,
title = {Human-in-the-loop machine learning: Reconceptualizing the role of the user in interactive approaches},
journal = {Internet of Things},
volume = {25},
pages = {101048},
year = {2024},
issn = {2542-6605},
doi = {https://doi.org/10.1016/j.iot.2023.101048},
url = {https://www.sciencedirect.com/science/article/pii/S2542660523003712},
author = {Oihane and Diego Casado-Mansilla and Diego López-de-Ipiña and Javier García-Zubia}
}

@inproceedings{buccinca2020proxy,title={Proxy tasks and subjective measures can be misleading in evaluating explainable AI systems},author={Bu{\c{c}}inca, Zana and others},booktitle={Proceedings of the 25th international conference on intelligent user interfaces},year={2020}}

@misc{zhou2020designing,
      title={Collaborative human-AI trust (CHAI-T): A process framework for active management of trust in human-AI collaboration}, 
      author={Melanie J. McGrath and Andreas Duenser and Justine Lacey and Cecile Paris},
      year={2024},
      eprint={2404.01615},
      archivePrefix={arXiv},
      primaryClass={cs.HC},
      url={https://arxiv.org/abs/2404.01615}, 
}

@misc{lee2022co,
      title={Human-AI Co-Creativity: Exploring Synergies Across Levels of Creative Collaboration}, 
      author={Jennifer Haase and Sebastian Pokutta},
      year={2024},
      eprint={2411.12527},
      archivePrefix={arXiv},
      primaryClass={cs.HC},
      url={https://arxiv.org/abs/2411.12527}, 
}

@article{kim2023socratic,
  title={SocraticLM: exploring socratic personalized teaching with large language models},
  author={Liu, Jiayu and Huang, Zhenya and Xiao, Tong and Sha, Jing and Wu, Jinze and Liu, Qi and Wang, Shijin and Chen, Enhong},
  journal={Advances in Neural Information Processing Systems},
  volume={37},
  pages={85693--85721},
  year={2024}
}

@article{krishna2022visual,title={Visual question generation: The state of the art},author={Patil, C. and Patwardhan, M.},journal={ACM Computing Surveys (CSUR)},volume={53},number={3},pages={1--22},year={2020},publisher={ACM New York, NY, USA}}

@misc{roberts2020measurement,
      title={Information Gain, Operator Spreading, and Sensitivity to Perturbations as Quantifiers of Chaos in Quantum Systems}, 
      author={Abinash Sahu},
      year={2024},
      eprint={2404.09464},
      archivePrefix={arXiv},
      primaryClass={quant-ph},
      url={https://arxiv.org/abs/2404.09464}, 
}

@misc{holstein2020participatory,
      title={Participatory AI: A Scandinavian Approach to Human-Centered AI}, 
      author={Niklas Elmqvist and Eve Hoggan and Hans-Jörg Schulz and Marianne Graves Petersen and Peter Dalsgaard and Ira Assent and Olav W. Bertelsen and Akhil Arora and Kaj Grønbæk and Susanne Bødker and Clemens Nylandsted Klokmose and Rachel Charlotte Smith and Sebastian Hubenschmid and Christoph A. Johns and Gabriela Molina León and Anton Wolter and Johannes Ellemose and Vaishali Dhanoa and Simon Aagaard Enni and Mille Skovhus Lunding and Karl-Emil Kjær Bilstrup and Juan Sánchez Esquivel and Luke Connelly and Rafael Pablos Sarabia and Morten Birk and Joachim Nyborg and Stefanie Zollmann and Tobias Langlotz and Meredith Siang-Yun Chou and Jens Emil Sloth Grønbæk and Michael Wessely and Yijing Jiang and Caroline Berger and Duosi Dai and Michael Mose Biskjaer and Germán Leiva and Jonas Frich and Eva Eriksson and Kim Halskov and Thorbjørn Mikkelsen and Nearchos Potamitis and Michel Yildirim and Arvind Srinivasan and Jeanette Falk and Nanna Inie and Ole Sejer Iversen and Hugo Andersson},
      year={2025},
      eprint={2509.12752},
      archivePrefix={arXiv},
      primaryClass={cs.HC},
      url={https://arxiv.org/abs/2509.12752}, 
}

@article{lee2022survey,title={Assessing human-AI interaction early through factorial surveys: a study on the guidelines for human-AI interaction},author={Li, Tianyi and others},journal={ACM Transactions on Computer-Human Interaction},volume={30},number={5},pages={1--45},year={2023},publisher={ACM New York, NY, USA}}

@misc{christiano2018supervising,
      title={Reversing the Paradigm: Building AI-First Systems with Human Guidance}, 
      author={Cosimo Spera and Garima Agrawal},
      year={2025},
      eprint={2506.12245},
      archivePrefix={arXiv},
      primaryClass={cs.AI},
      url={https://arxiv.org/abs/2506.12245}, 
}

@book{cover2006elements,
  title={Elements of Information Theory},
  author={Cover, Thomas M and Thomas, Joy A},
  year={2006},
  publisher={Wiley}
}

@misc{lambert2022illustrating,
      title={Reincarnating Reinforcement Learning: Reusing Prior Computation to Accelerate Progress}, 
      author={Rishabh Agarwal and Max Schwarzer and Pablo Samuel Castro and Aaron Courville and Marc G. Bellemare},
      year={2022},
      eprint={2206.01626},
      archivePrefix={arXiv},
      primaryClass={cs.LG},
      url={https://arxiv.org/abs/2206.01626}, 
}

@article{green2020using,
  author={Basole, Rahul C. and Major, Timothy},
  journal={IEEE Computer Graphics and Applications}, 
  title={Generative AI for Visualization: Opportunities and Challenges}, 
  year={2024},
  volume={44},
  number={2},
  pages={55-64},
  doi={10.1109/MCG.2024.3362168}}

@misc{liu2023chartllm,
      title={ChartLlama: A Multimodal LLM for Chart Understanding and Generation}, 
      author={Yucheng Han and Chi Zhang and Xin Chen and Xu Yang and Zhibin Wang and Gang Yu and Bin Fu and Hanwang Zhang},
      year={2023},
      eprint={2311.16483},
      archivePrefix={arXiv},
      primaryClass={cs.CV},
      url={https://arxiv.org/abs/2311.16483}, 
}

@misc{zhou2023survey,
      title={A Systematic Review of Human-AI Co-Creativity}, 
      author={Saloni Singh and Koen Hindriks and Dirk Heylen and Kim Baraka},
      year={2025},
      eprint={2506.21333},
      archivePrefix={arXiv},
      primaryClass={cs.HC},
      url={https://arxiv.org/abs/2506.21333}, 
}

@article{young2013pomdp,
  title={POMDP-based statistical spoken dialog systems: A review},
  author={Young, Steve and Ga{\v{s}}i{\'c}, Milica and Thomson, Blaise and Williams, Jason D},
  journal={Proceedings of the IEEE},
  volume={101},
  number={5},
  pages={1160--1179},
  year={2013}
}

@inproceedings{wen2017network,
  title={A network-based end-to-end trainable task-oriented dialogue system},
  author={Wen, Tsung-Hsien and Vandyke, David and Mrksic, Nikola and Gasic, Milica and Rojas-Barahona, Lina M and Su, Pei-Hao and Ultes, Stefan and Young, Steve},
  booktitle={EACL},
  year={2017}
}

@inproceedings{budzianowski2018multiwoz,
  title={MultiWOZ--A Large-Scale Multi-Domain Wizard-of-Oz Dataset for Task-Oriented Dialogue Modelling},
  author={Budzianowski, Pawe{\l} and Wen, Tsung-Hsien and Tseng, Bo-Hsiang and Casanueva, I{\~n}igo and Ultes, Stefan and Ramadan, Osman and Gasic, Milica},
  booktitle={EMNLP},
  year={2018}
}

@misc{xu2024futureclarification,
      title={Modeling Future Conversation Turns to Teach LLMs to Ask Clarifying Questions}, 
      author={Michael J. Q. Zhang and W. Bradley Knox and Eunsol Choi},
      year={2025},
      eprint={2410.13788},
      archivePrefix={arXiv},
      primaryClass={cs.CL},
      url={https://arxiv.org/abs/2410.13788}, 
}

@inproceedings{woz1,
    Author = {Budzianowski, Pawe{\l} and Wen, Tsung-Hsien and Tseng, Bo-Hsiang  and Casanueva, I{\~n}igo and Ultes Stefan and Ramadan Osman and Ga{\v{s}}i\'c, Milica},
    title={MultiWOZ - A Large-Scale Multi-Domain Wizard-of-Oz Dataset for Task-Oriented Dialogue Modelling},
    booktitle={Proceedings of the 2018 Conference on Empirical Methods in Natural Language Processing (EMNLP)},
    year={2018}
}

@inproceedings{woz2,
  title={Large-Scale Multi-Domain Belief Tracking with Knowledge Sharing},
  author={Ramadan, Osman and Budzianowski, Pawe{\l} and Gasic, Milica},
  booktitle={Proceedings of the 56th Annual Meeting of the Association for Computational Linguistics},
  volume={2},
  pages={432--437},
  year={2018}
}

@article{woz3,
  title={MultiWOZ 2.1: Multi-Domain Dialogue State Corrections and State Tracking Baselines},
  author={Eric, Mihail and Goel, Rahul and Paul, Shachi and Sethi, Abhishek and Agarwal, Sanchit and Gao, Shuyag and Hakkani-Tur, Dilek},
  journal={arXiv preprint arXiv:1907.01669},
  year={2019}
}

@inproceedings{woz4,
  title={MultiWOZ 2.2: A Dialogue Dataset with Additional Annotation Corrections and State Tracking Baselines},
  author={Zang, Xiaoxue and Rastogi, Abhinav and Sunkara, Srinivas and Gupta, Raghav and Zhang, Jianguo and Chen, Jindong},
  booktitle={Proceedings of the 2nd Workshop on Natural Language Processing for Conversational AI, ACL 2020},
  pages={109--117},
  year={2020}
}

@article{settles2009active,
  title={The Power of Ensembles for Active Learning in Image Classification},
  author={William H. Beluch and Tim Genewein and A. N{\"u}rnberger and Jan M. K{\"o}hler},
  journal={2018 IEEE/CVF Conference on Computer Vision and Pattern Recognition},
  year={2018},
  pages={9368-9377},
  url={https://api.semanticscholar.org/CorpusID:52838058}
}

@article{lewis1994sequential,
  title={A sequential algorithm for training text classifiers},
  author={Lewis, David D and Gale, William A},
  journal={SIGIR},
  year={1994}
}

@misc{zhang2021dialogpolicy,
      title={Dialog Policy Learning for Joint Clarification and Active Learning Queries}, 
      author={Aishwarya Padmakumar and Raymond J. Mooney},
      year={2020},
      eprint={2006.05456},
      archivePrefix={arXiv},
      primaryClass={cs.CV},
      url={https://arxiv.org/abs/2006.05456}, 
}

@inproceedings{pan2024informativequestions,
    title = "Learning to Ask Informative Questions: Enhancing {LLM}s with Preference Optimization and Expected Information Gain",
    author = "Mazzaccara, Davide  and
      Testoni, Alberto  and
      Bernardi, Raffaella",
    editor = "Al-Onaizan, Yaser  and
      Bansal, Mohit  and
      Chen, Yun-Nung",
    booktitle = "Findings of the Association for Computational Linguistics: EMNLP 2024",
    month = nov,
    year = "2024",
    address = "Miami, Florida, USA",
    publisher = "Association for Computational Linguistics",
    url = "https://aclanthology.org/2024.findings-emnlp.291/",
    doi = "10.18653/v1/2024.findings-emnlp.291",
    pages = "5064--5074"
}

@inproceedings{xing2024,
    title = "Learning to Ask Denotative and Connotative Questions for Knowledge-based {VQA}",
    author = "Xing, Xiaoying  and
      Xiong, Peixi  and
      Fan, Lei  and
      Li, Yunxuan  and
      Wu, Ying",
    editor = "Al-Onaizan, Yaser  and
      Bansal, Mohit  and
      Chen, Yun-Nung",
    booktitle = "Findings of the Association for Computational Linguistics: EMNLP 2024",
    month = nov,
    year = "2024",
    address = "Miami, Florida, USA",
    publisher = "Association for Computational Linguistics",
    url = "https://aclanthology.org/2024.findings-emnlp.487/",
    doi = "10.18653/v1/2024.findings-emnlp.487",
    pages = "8301--8315"
}

@misc{hu2024uncertainty,
      title={Uncertainty of Thoughts: Uncertainty-Aware Planning Enhances Information Seeking in Large Language Models}, 
      author={Zhiyuan Hu and Chumin Liu and Xidong Feng and Yilun Zhao and See-Kiong Ng and Anh Tuan Luu and Junxian He and Pang Wei Koh and Bryan Hooi},
      year={2024},
      eprint={2402.03271},
      archivePrefix={arXiv},
      primaryClass={cs.CL},
      url={https://arxiv.org/abs/2402.03271}, 
}

@inproceedings{takanobu-etal-2019-guided,
    title = "Guided Dialog Policy Learning: Reward Estimation for Multi-Domain Task-Oriented Dialog",
    author = "Takanobu, Ryuichi  and
      Zhu, Hanlin  and
      Huang, Minlie",
    editor = "Inui, Kentaro  and
      Jiang, Jing  and
      Ng, Vincent  and
      Wan, Xiaojun",
    booktitle = "Proceedings of the 2019 Conference on Empirical Methods in Natural Language Processing and the 9th International Joint Conference on Natural Language Processing (EMNLP-IJCNLP)",
    month = nov,
    year = "2019",
    address = "Hong Kong, China",
    publisher = "Association for Computational Linguistics",
    url = "https://aclanthology.org/D19-1010/",
    doi = "10.18653/v1/D19-1010",
    pages = "100--110"
}

@inproceedings{christiano2017deep,
  title={Deep reinforcement learning from human preferences},
  author={Christiano, Paul and Leike, Jan and Brown, Tom and Martic, Miljan and Legg, Shane and Amodei, Dario},
  booktitle={NeurIPS},
  year={2017}
}

@article{ouyang2022training,
  title={Training language models to follow instructions with human feedback},
  author={Ouyang, Long and et al.},
  journal={arXiv preprint arXiv:2203.02155},
  year={2022}
}

@inproceedings{schulman2017ppo,
  title={Proximal Policy Optimization Algorithms},
  author={Schulman, John and Wolski, Filip and Dhariwal, Prafulla and Radford, Alec and Klimov, Oleg},
  booktitle={arXiv preprint arXiv:1707.06347},
  year={2017}
}

@misc{rafailov2023grpo,
      title={DeepSeekMath: Pushing the Limits of Mathematical Reasoning in Open Language Models}, 
      author={Zhihong Shao and Peiyi Wang and Qihao Zhu and Runxin Xu and Junxiao Song and Xiao Bi and Haowei Zhang and Mingchuan Zhang and Y. K. Li and Y. Wu and Daya Guo},
      year={2024},
      eprint={2402.03300},
      archivePrefix={arXiv},
      primaryClass={cs.CL},
      url={https://arxiv.org/abs/2402.03300}, 
}

@inproceedings{rafailov2023dpo,
  title={Direct Preference Optimization: Your Language Model is Secretly a Reward Model},
  author={Rafailov, Rafael and others},
  booktitle={NeurIPS},
  year={2023}
}

@misc{bai2022constitutional,
      title={Constitutional AI: Harmlessness from AI Feedback}, 
      author={Yuntao Bai and Saurav Kadavath and Sandipan Kundu and Amanda Askell and Jackson Kernion and Andy Jones and Anna Chen and Anna Goldie and Azalia Mirhoseini and Cameron McKinnon and Carol Chen and Catherine Olsson and Christopher Olah and Danny Hernandez and Dawn Drain and Deep Ganguli and Dustin Li and Eli Tran-Johnson and Ethan Perez and Jamie Kerr and Jared Mueller and Jeffrey Ladish and Joshua Landau and Kamal Ndousse and Kamile Lukosuite and Liane Lovitt and Michael Sellitto and Nelson Elhage and Nicholas Schiefer and Noemi Mercado and Nova DasSarma and Robert Lasenby and Robin Larson and Sam Ringer and Scott Johnston and Shauna Kravec and Sheer El Showk and Stanislav Fort and Tamera Lanham and Timothy Telleen-Lawton and Tom Conerly and Tom Henighan and Tristan Hume and Samuel R. Bowman and Zac Hatfield-Dodds and Ben Mann and Dario Amodei and Nicholas Joseph and Sam McCandlish and Tom Brown and Jared Kaplan},
      year={2022},
      eprint={2212.08073},
      archivePrefix={arXiv},
      primaryClass={cs.CL},
      url={https://arxiv.org/abs/2212.08073}, 
}

@article{levine2020offline,
  title={Offline reinforcement learning: Tutorial, review, and perspectives on open problems},
  author={Levine, Sergey and Kumar, Aviral and Tucker, George and Fu, Justin},
  journal={arXiv preprint arXiv:2005.01643},
  year={2020}
}

@inproceedings{xiao-etal-2025-infopo,
  title = {InfoPO: On Mutual Information Maximization for Large Language Model Alignment},
  author = {Xiao, Teng and Ge, Zhen and Sanghavi, Sujay and Wang, Tian and Katz-Samuels, Julian and Versage, Marc and Cui, Qingjun and Chilimbi, Trishul},
  booktitle = {Proceedings of the 2025 NAACL: Human Language Technologies, Long Papers},
  pages = {11699--11711},
  year = {2025},
  month = {April},
  address = {Albuquerque, New Mexico},
  publisher = {Association for Computational Linguistics}
}

@article{ilkou2025hybrid,
  title = {Hybrid Evaluation of Socratic Dialogue for Teaching},
  author = {Ilkou, Eleni and Linzbach, Stephan and Wallat, Jonas},
  journal = {CEUR Workshop Proceedings},
  volume = {3953},
  year = {2025}
}

@misc{kostrikov2022offline,
      title={Offline Reinforcement Learning with Implicit Q-Learning}, 
      author={Ilya Kostrikov and Ashvin Nair and Sergey Levine},
      year={2021},
      eprint={2110.06169},
      archivePrefix={arXiv},
      primaryClass={cs.LG},
      url={https://arxiv.org/abs/2110.06169}, 
}

@inproceedings{cq-gen,
    title = "Asking Clarification Questions to Handle Ambiguity in Open-Domain {QA}",
    author = "Lee, Dongryeol  and
      Kim, Segwang  and
      Lee, Minwoo  and
      Lee, Hwanhee  and
      Park, Joonsuk  and
      Lee, Sang-Woo  and
      Jung, Kyomin",
    editor = "Bouamor, Houda  and
      Pino, Juan  and
      Bali, Kalika",
    booktitle = "Findings of the Association for Computational Linguistics: EMNLP 2023",
    month = dec,
    year = "2023",
    address = "Singapore",
    publisher = "Association for Computational Linguistics",
    url = "https://aclanthology.org/2023.findings-emnlp.772/",
    doi = "10.18653/v1/2023.findings-emnlp.772",
    pages = "11526--11544"
}

@misc{huang2025codeclarify,
      title={Curiosity by Design: An LLM-based Coding Assistant Asking Clarification Questions}, 
      author={Harsh Darji and Thibaud Lutellier},
      year={2025},
      eprint={2507.21285},
      archivePrefix={arXiv},
      primaryClass={cs.AI},
      url={https://arxiv.org/abs/2507.21285}, 
}

@misc{lee2024rlaif,title={Rlaif: Scaling reinforcement learning from human feedback with ai feedback},author={Lee, H. and Phatale, S. and Mansoor, H. and Lu, K. R. and Mesnard, T. and Ferret, J. and others and Rastogi, A.},year={2023}}

@misc{entropy_2,
      title={CoCoA: Confidence and Context-Aware Adaptive Decoding for Resolving Knowledge Conflicts in Large Language Models}, 
      author={Anant Khandelwal and Manish Gupta and Puneet Agrawal},
      year={2025},
      eprint={2508.17670},
      archivePrefix={arXiv},
      primaryClass={cs.CL},
      url={https://arxiv.org/abs/2508.17670}, 
}

@misc{entropy_1,
      title={Active Learning with Context Sampling and One-vs-Rest Entropy for Semantic Segmentation}, 
      author={Fei Wu and Pablo Marquez-Neila and Hedyeh Rafi-Tarii and Raphael Sznitman},
      year={2025},
      eprint={2412.06470},
      archivePrefix={arXiv},
      primaryClass={cs.CV},
      url={https://arxiv.org/abs/2412.06470}, 
}

@misc{socratic,
      title={Prompting Large Language Models With the Socratic Method}, 
      author={Edward Y. Chang},
      year={2023},
      eprint={2303.08769},
      archivePrefix={arXiv},
      primaryClass={cs.LG},
      url={https://arxiv.org/abs/2303.08769}, 
}

@misc{activate_1,
      title={ALFred: An Active Learning Framework for Real-world Semi-supervised Anomaly Detection with Adaptive Thresholds}, 
      author={Shanle Yao and Ghazal Alinezhad Noghre and Armin Danesh Pazho and Hamed Tabkhi},
      year={2025},
      eprint={2508.09058},
      archivePrefix={arXiv},
      primaryClass={cs.CV},
      url={https://arxiv.org/abs/2508.09058}, 
}

@misc{activate_2,
      title={OFAL: An Oracle-Free Active Learning Framework}, 
      author={Hadi Khorsand and Vahid Pourahmadi},
      year={2025},
      eprint={2508.08126},
      archivePrefix={arXiv},
      primaryClass={cs.LG},
      url={https://arxiv.org/abs/2508.08126}, 
}

@misc{activate_3,
      title={Improving Active Learning with a Bayesian Representation of Epistemic Uncertainty}, 
      author={Jake Thomas and Jeremie Houssineau},
      year={2024},
      eprint={2412.08225},
      archivePrefix={arXiv},
      primaryClass={stat.ME},
      url={https://arxiv.org/abs/2412.08225}, 
}

@inproceedings{geishauser2021what,
  title={What Does The User Want? Information Gain for Hierarchical Dialogue Policy Optimisation},
  author={Geishauser, Christian and Hu, Songbo and Lin, Hsien-chin and Lubis, Nurul and Heck, Michael and Feng, Shutong and van Niekerk, Carel and Gašić, Milica},
  booktitle={Proceedings of the 22nd Annual Meeting of the Special Interest Group on Discourse and Dialogue (SIGDIAL)},
  year={2021},
  url={https://arxiv.org/abs/2109.07129}
}

@misc{active_pref_inference2023,
      title={Active Preference Inference using Language Models and Probabilistic Reasoning}, 
      author={Wasu Top Piriyakulkij and Volodymyr Kuleshov and Kevin Ellis},
      year={2024},
      eprint={2312.12009},
      archivePrefix={arXiv},
      primaryClass={cs.CL},
      url={https://arxiv.org/abs/2312.12009}, 
}

@article{seed_grpo2025,
  title={SEED-GRPO: Semantic Entropy Enhanced GRPO for Uncertainty-Aware Policy Optimization},
  author={Chen, Minghan and Chen, Guikun and Wang, Wenguan and Yang, Yi},
  journal={arXiv preprint arXiv:2505.12346},
  year={2025},
  url={https://arxiv.org/abs/2505.12346}
}

@misc{dynamic_profile_modeling2025,
      title={Teaching Language Models to Evolve with Users: Dynamic Profile Modeling for Personalized Alignment}, 
      author={Weixiang Zhao and Xingyu Sui and Yulin Hu and Jiahe Guo and Haixiao Liu and Biye Li and Yanyan Zhao and Bing Qin and Ting Liu},
      year={2025},
      eprint={2505.15456},
      archivePrefix={arXiv},
      primaryClass={cs.CL},
      url={https://arxiv.org/abs/2505.15456}, 
}

@article{d_2,
author = {Joshua B. Tenenbaum  and Charles Kemp  and Thomas L. Griffiths  and Noah D. Goodman },
title = {How to Grow a Mind: Statistics, Structure, and Abstraction},
journal = {Science},
volume = {331},
number = {6022},
pages = {1279-1285},
year = {2011},
doi = {10.1126/science.1192788},
URL = {https://www.science.org/doi/abs/10.1126/science.1192788},
eprint = {https://www.science.org/doi/pdf/10.1126/science.1192788}}

@inproceedings{d_3,
  title={An asymptotic test for conditional independence using analytic kernel embeddings},
  author={Scetbon, Meyer and Meunier, Laurent and Romano, Yaniv},
  booktitle={International Conference on Machine Learning},
  pages={19328--19346},
  year={2022},
  organization={PMLR}
}

@article{d_4,
  title={Conditional independence testing with heteroskedastic data and applications to causal discovery},
  author={G{\"u}nther, Wiebke and Ninad, Urmi and Wahl, Jonas and Runge, Jakob},
  journal={Advances in Neural Information Processing Systems},
  volume={35},
  pages={16191--16202},
  year={2022}
}

@inproceedings{d_5,
author = {Polo, Felipe Maia and Sun, Yuekai and Banerjee, Moulinath},
title = {Conditional independence testing under misspecified inductive biases},
year = {2023},
publisher = {Curran Associates Inc.},
address = {Red Hook, NY, USA},
booktitle = {Proceedings of the 37th International Conference on Neural Information Processing Systems},
articleno = {2553},
numpages = {36},
location = {New Orleans, LA, USA},
series = {NIPS '23}
}

@InProceedings{d_6,
  title = 	 {Scalable Causal Structure Learning via Amortized Conditional Independence Testing},
  author =       {Leiner, James and Manzo, Brian and Ramdas, Aaditya and Tansey, Wesley},
  booktitle = 	 {Proceedings of the Fourth Conference on Causal Learning and Reasoning},
  pages = 	 {174--200},
  year = 	 {2025},
  editor = 	 {Huang, Biwei and Drton, Mathias},
  volume = 	 {275},
  series = 	 {Proceedings of Machine Learning Research},
  month = 	 {07--09 May},
  publisher =    {PMLR},
  url = 	 {https://proceedings.mlr.press/v275/leiner25a.html}
}

@inproceedings{d_7,
 author = {Bellot, Alexis and van der Schaar, Mihaela},
 booktitle = {Advances in Neural Information Processing Systems},
 editor = {H. Wallach and H. Larochelle and A. Beygelzimer and F. d\textquotesingle Alch\'{e}-Buc and E. Fox and R. Garnett},
 pages = {},
 publisher = {Curran Associates, Inc.},
 title = {Conditional Independence Testing using Generative Adversarial Networks},
 url = {https://proceedings.neurips.cc/paper_files/paper/2019/file/dc87c13749315c7217cdc4ac692e704c-Paper.pdf},
 volume = {32},
 year = {2019}
}

@inproceedings{d_8,
    title = "A Conditional Variational Framework for Dialog Generation",
    author = "Shen, Xiaoyu  and
      Su, Hui  and
      Li, Yanran  and
      Li, Wenjie  and
      Niu, Shuzi  and
      Zhao, Yang  and
      Aizawa, Akiko  and
      Long, Guoping",
    editor = "Barzilay, Regina  and
      Kan, Min-Yen",
    booktitle = "Proceedings of the 55th Annual Meeting of the Association for Computational Linguistics (Volume 2: Short Papers)",
    month = jul,
    year = "2017",
    address = "Vancouver, Canada",
    publisher = "Association for Computational Linguistics",
    url = "https://aclanthology.org/P17-2080/",
    doi = "10.18653/v1/P17-2080",
    pages = "504--509"
}

@inproceedings{su_1,
    title = "The Art of {SOCRATIC} {QUESTIONING}: Recursive Thinking with Large Language Models",
    author = "Qi, Jingyuan  and
      Xu, Zhiyang  and
      Shen, Ying  and
      Liu, Minqian  and
      Jin, Di  and
      Wang, Qifan  and
      Huang, Lifu",
    editor = "Bouamor, Houda  and
      Pino, Juan  and
      Bali, Kalika",
    booktitle = "Proceedings of the 2023 Conference on Empirical Methods in Natural Language Processing",
    month = dec,
    year = "2023",
    address = "Singapore",
    publisher = "Association for Computational Linguistics",
    url = "https://aclanthology.org/2023.emnlp-main.255/",
    doi = "10.18653/v1/2023.emnlp-main.255",
    pages = "4177--4199"
}
\bibliographystyle{icml2026}

\newpage
\appendix
\onecolumn

\section{Appendix}

\appendix

\vspace{-1mm}
\section{Extended Discussion of Related Work}
\label{sec:Related Work}
\vspace{-1mm}
\textbf{Clarification and Inquiry as Strategy.}
A growing body of work recognizes the strategic value of asking clarifying questions. In open-domain QA, for example, clarification has been shown to improve accuracy by resolving ambiguity before answering~\citep{cq-gen}. Other approaches model the decision of whether and when to ask a question based on the expected utility of future dialogue turns, effectively learning an optimal timing policy~\citep{xu2024futureclarification}. In specialized domains like code generation, clarification also improves correctness, highlighting its broad value~\citep{huang2025codeclarify}. While these methods validate the importance of proactive inquiry, they often optimize for single-answer correctness using heuristic signals or rely on downstream annotations to estimate future value. Nous shifts the focus from when to ask to what to ask. Our framework aims for convergence toward a complete, high-dimensional specification, where the reward is an immediate, intrinsic signal derived from entropy reduction over structured attributes, providing a stable, cumulative signal for optimizing the content of each inquiry.

\textbf{Information Gain as a Measure of Question Quality.}
Our work builds on the principle of using information theory to quantify question value. In task-oriented dialogue, early frameworks used reward estimation to guide policy learning, though often as a proxy for external goals like booking success~\citep{takanobu-etal-2019-guided, geishauser2021what}. More directly, work in visual dialogue has used information gain to explicitly model the value of ``confirmation questions" (\textit{e.g.}, yes/no questions), demonstrating that such inquiries efficiently reduce the candidate set and improve success rates in guessing games like GuessWhat?!~\citep{hu2024uncertainty}. Similarly, recent research establishes the ``20 questions" game as a benchmark for active information seeking in LLMs, using expected information gain to rank and select the most discriminative question from a set of candidates generated via Chain-of-Thought prompting~\citep{pan2024informativequestions, roberts2020measurement}. 
These studies collectively affirm that an entropy-based objective is a powerful tool for guiding efficient inquiry.
However, directly applying existing information-theoretic methods to open-ended construction tasks faces significant challenges. For instance, UoT (Uncertainty of Thoughts)~\citep{hu2024uncertainty} relies on simulation-based planning, which becomes computationally intractable in our high-dimensional state space ($>35,000$ combinations) due to the curse of dimensionality. Similarly, prompting-based clarification methods like CQ-Gen~\citep{cq-gen} often focus on high-level semantic disambiguation rather than structural constraint resolution. In contrast, Nous integrates entropy reduction as a real-time intrinsic reward for a generative policy, avoiding the need for expensive full enumeration while maintaining precision in structural alignment.
Nous integrates and advances these ideas into a new learning paradigm. Instead of using information gain as a post-hoc selection heuristic~\citep{xiao-etal-2025-infopo} or applying it to a constrained set of question types, we use it as a real-time, intrinsic reward to train a generative policy. This enables Nous to learn to generate open-ended, natural language questions, which offers a significant advantage in high-dimensional, structured design spaces. In this way, we bridge the gap between the theoretical appeal of information gain and the practical challenge of training a proactive conversational agent for complex, creative tasks.

\textbf{Socratic Prompting versus Learnable Strategy.}
Socratic prompting, exemplified by \textit{Prompting Large Language Models with the Socratic Method}~\citep{socratic}, encourages models to ask questions before answering through templates. \textit{SocraticLM: Exploring Socratic Personalized Teaching}~\citep{kim2023socratic} extends this to personalized instruction, while \textit{Hybrid Evaluation of Socratic Dialogue for Teaching}~\citep{ilkou2025hybrid} evaluates its educational benefits and limits. While these approaches highlight the pedagogical value of Socratic interaction, they remain prompt-based. Nous extends the paradigm into a trainable policy: information gain defines the objective, and offline preference data enables optimization. This transforms “asking questions” from prompt-driven behavior into a generalizable capability robust across user types.

\textbf{Comparison with Traditional Slot-Filling Paradigms.} While inspired by traditional Task-Oriented Dialogue (TOD)\citep{young2013pomdp, wen2017network}, Nous addresses a fundamentally different problem scope. TOD systems typically perform extraction and retrieval: the user's intent is assumed to map to a specific entry in a database (\textit{e.g.}, a restaurant), and the system's goal is to fill static slots to filter this database\citep{budzianowski2018multiwoz}. Nous, conversely, performs clarification and construction: the target (a scientific diagram) does not yet exist. In our scenario, the system does not have a fixed list of questions or slots. Its core task is to enable the model to proactively detect and eliminate all uncertainties in the user's description before performing the generation operation\citep{pan2024informativequestions}. This strategy of ``thoroughly clarifying first, then precisely executing" allows the model to spontaneously adjust its inquiry strategy based on the degree of ambiguity in the user's intent\citep{christiano2017deep}. This not only effectively aligns with the vague ideas in the user's mind but also fundamentally avoids the subsequent tedious trial-and-error and repeated modification processes. It is difficult to achieve this with traditional TOD methods.

\textbf{Summary.}
Prior work can be grouped into two broad directions: Socratic prompting methods that encourage proactive questioning through templates or pedagogy, and a method to achieve accurate question answering by quantifying the value of questions through entropy or mutual information.
Nous advances both threads by combining structured belief states, closed-form entropy-based rewards, and offline policy optimization, thereby making clarification a new perspective for bridging the intention gap, rather than a heuristic method or template.

\textbf{Horizontal comparison experiment.}
\label{sec:horizontal}
To further demonstrate the superiority of our method, we selected two methods with the most similar functionalities, namely UoT (NeurIPS 2024)~\citep{hu2024uncertainty} and CQ-Gen (EMNLP 2023)~\citep{cq-gen}. Since these methods cannot output the final summary text description (lacking the ability to generate standardized and complete content), we only compared the interaction efficiency and the total cumulative information. We adopted two experimental settings, divided into (A) zero-start group and (B) draft-start group. The experimental results are shown in Table~\ref{tab:Supplementary_horizontal_comparison_experiments}.

UoT (NeurIPS 2024): Since full enumeration is infeasible, we limited it to sparse sampling (N=5). The results show that due to inaccurate valuation, the model fails to pose effective questions, leading to a surge in interaction rounds (an average of 35.6 rounds) and low efficiency in information acquisition. CQ-Gen (EMNLP 2023): This model tends to ask high-level semantic questions such as ``purpose" and cannot track structural parameters. Although the dialogue is short, the total cumulative information is extremely low (only 12.2 on average), indicating that it cannot complete the construction of structural information. It is worth noting that since these two methods cannot generate summary descriptions and are not designed for chart generation, we cannot compare the quality of the final generated content, and can only make a simple horizontal comparison from the efficiency of information acquisition.

\begin{table*}[!ht]
\centering
\caption{Supplementary horizontal comparison experiments. As baseline models (UoT, CQ-Gen) cannot generate the finalized structural prompt for image generation, we compare only the interaction efficiency (Dialogue Turns) and the total Information Gain (Total IG). \textbf{Nous (OfG)} denotes our proposed method.}
\label{tab:Supplementary_horizontal_comparison_experiments}
\setlength{\tabcolsep}{5pt} 
\begin{tabular}{llcc} 
\toprule
Prompt Type & Method & Turns(↓) & Total IG(↑) \\ 

\midrule
& Nous(OfG) & 17.3 & 126.44 \\ 
Promp\_Zero & UoT    & 33.6  & 38.38 \\ 
& CQ-Gen    & 7.1 & 9.72 \\ 
\midrule
& Nous(OfG) & 7.6 & 42.31 \\ 
Prompt\_Draft & UoT & 37.6 & 12.30 \\ 
& CQ-Gen    & 9.7 & 14.61 \\ 
\bottomrule
\end{tabular}
\vspace{-2mm}
\end{table*}

\section{Supplementary to Ablation Experiments}


To investigate the impact of training data quality on the final policy, we adjusted the data generation process. In addition to the Template Oracle, we introduced two variants: a Vague Oracle (providing incomplete information) and a Noisy Oracle (interjecting irrelevant information in its responses). Using these three data sources, we trained three respective models: Nous-Template, Nous-Vague, and Nous-Noisy. Distinct from the discussion on user expertise in Section \ref{sec:User Expertise}, this section evaluates our framework's learning ability under different training data conditions.

The experimental results, shown in Table \ref{tab:Supplementary_ablation_expertise}, reveal the following:
\textbf{Adaptability to Vagueness.} Nous-Vague's performance in standard tests was comparable to the baseline model. This demonstrates the framework's effectiveness: although its training data (19,123 samples) was longer than the baseline data (11,851 samples) due to more clarification turns, leading to increased training time, the model still learned the core strategy of identifying high information-gain questions from these seemingly ``inefficient" dialogues.
\textbf{Filtering of Noise.} Nous-Noisy also performed nearly identically to the baseline model. This reveals a key property of our information-theoretic reward: it has a natural ``immunity" to irrelevant information. Since noise cannot reduce the entropy of any attribute, its information gain reward is zero. Consequently, the training process automatically filters out the impact of noise, allowing the model to focus on learning genuinely effective question-answer patterns.

This study demonstrates our framework's high robustness to training data quality. Crucially, it also validates the robustness of our semantic parser, which successfully maps varied and imperfect responses back to the same underlying attributes, a key requirement for real-world application.

\begin{table*}[!ht]
\centering
\small
\caption{Supplementary ablation study experimental results. Considering the significant time costs associated with data construction, model training, and drawing using VisPainter, the evaluation results of VisPainter are omitted in this experiment. Conduct model image generation  evaluation experiments using nano-banana.}
\label{tab:Supplementary_ablation_expertise}
\setlength{\tabcolsep}{7pt} 
\begin{tabular}{lcccccccc}
\toprule
& Turns & Total IG  & \multicolumn{3}{c}{Human Judge(↑)} & \multicolumn{3}{c}{GPT-5 Judge(↑)} \\
\cmidrule(lr){4-6} \cmidrule(lr){7-9}
Method & (↓) & (↑)  &\footnotesize \textbf{Win} & \footnotesize \textbf{W/T(0.5)} & \footnotesize \textbf{W/T} &\footnotesize \textbf{Win} & \footnotesize \textbf{W/T(0.5)} & \footnotesize \textbf{W/T}  \\
\midrule
Nous-Template & 20.3 & \textbf{120.5}  & 0.29 & \textbf{0.53} & \textbf{0.76} & \textbf{0.34} & \textbf{0.52} & 0.69 \\
Nous-Vague    & 22.1 & 117.1  & \textbf{0.30} & 0.50 & 0.69 & 0.29 & 0.50 & \textbf{0.70} \\
Nous-Noisy    & \textbf{19.7} & 115.8  & 0.26 & 0.48 & 0.70 & 0.33 & 0.49 & 0.65 \\
\bottomrule
\end{tabular}
\vspace{-2mm}
\end{table*}

\section{Exploration of Generalization Potential}
\label{sec:Applicability Boundaries}

\vspace{-2mm}
\paragraph{Experimental Setting.}
To test whether our framework has the potential to be generalized beyond scientific diagram generation, we evaluated it in collaborative novel writing. This domain differs substantially from diagram creation in both task structure and interaction dynamics, yet retains properties that make systematic study feasible. Novel writing is open-ended and creative, but it is also composed of recurring elements such as characters, settings, and events. These elements can be represented as structured state vectors, enabling the construction of a empirical prior and the computation of per-turn information gain. At the same time, evaluation is relatively tractable: the quality of co-created narratives can be assessed through outline coverage and comparative judgments of readability and fidelity. These characteristics make collaborative novel writing another ideal testbed for examining the generality of our Socratic inquiry framework.

\vspace{-2mm}
\paragraph{Data Preparation and Training.}

We collect novels from publicly available corpora. Since long-form narratives are often lengthy and would substantially increase the workload, we simplify the data by selecting representative chapters as test material, which are further rewritten through AI-assisted editing to avoid copyright concerns. In total, we obtain 120 processed samples, with 100 used for training and 20 for testing. From each sample, we extract structured elements such as characters, settings, conflicts, and resolutions to form state vectors and construct a empirical prior as the prior. The data construction process follows the main text: the ground truth outline is provided to an Oracle, which answers model queries during simulation. Each question–answer pair is scored by information gain to create a preference dataset. Nous (OfG) is trained with offline GRPO, Nous (SFT) with supervised fine-tuning, and GPT baselines (zero-shot and few-shot: GPT-zero, GPT-fews) are included for evaluation.

\vspace{-2mm}

\paragraph{Evaluation Metrics.}
For evaluation, we adopt two dimensions consistent with the main paper: interaction efficiency and output quality. Interaction efficiency is measured by dialogue turns and total information gain, reflecting whether a model can ask high quality questions within a limited number of turns. Output quality is assessed through outline coverage and subjective quality evaluation. Specifically, we compare the generated summaries of novel passages using both human and GPT judges in pairwise evaluations. These metrics provide a balanced view of how effectively the models gather information and how well they translate it into coherent creative output.

\vspace{-2mm}
\paragraph{Results and Discussion.} 
Novel writing represents a common and relatively structured domain, where LLMs already possess strong intrinsic capabilities. As shown in Table~\ref{tab:novel_generalization}, this leads to notable efficiency for untrained models, which complete dialogues in fewer turns. However, Nous (OfG) achieves about 15\% higher cumulative information gain compared to untrained baselines, confirming the benefit of entropy-based training. In terms of outline coverage, both OfG and SFT perform strongly, while GPT-few and GPT-zero show little distinction. In subjective evaluations by human and GPT-5 judges, the trained models consistently outperformed the baseline models, and this experimental result is consistent with the conclusions in our main text. Overall, these results verify that our framework remains effective in a distinctly different creative domain, demonstrate the generalization potential of our method, and provide insights for future research directions in broader applications.

\paragraph{Applicability Boundaries and Marginal Utility.} While the experiments confirm the mechanism's universality, we observe a difference in the magnitude of improvement between the scientific diagram task (main paper) and the novel writing task. We attribute this difference primarily to the \textbf{High Baseline Effect}. Modern LLMs have internalized massive amounts of narrative structures during pre-training, providing them with a strong prior for storytelling. Even without active inquiry, baselines like GPT-4 can generate coherent narratives, leading to diminishing marginal returns for additional clarification. In contrast, scientific diagramming is an atypical generation task requiring precise spatial logic and strict constraints-areas where LLM priors are weak. Consequently, Nous delivers a qualitative leap in the diagram domain, whereas in the novel domain, it provides incremental optimization.

\textbf{Definition of applicable boundaries:} Based on the above findings, we further clarify the ``optimal application area" of ours framework. We posit that the premise for Nous to demonstrate excellent effectiveness needs to meet two core characteristics: First, \textbf{high structural constraints}: Tasks should have an objective logical or topological structure (such as scientific flowcharts) rather than random divergent thinking. Only within a deterministic logical space can the system accurately calculate information entropy values and ask efficient questions. Second, \textbf{significant intention gap}: where the user has a specific and complex goal but finds it difficult to express it clearly. If a task allows for arbitrary open-ended generation and the user's intention itself is divergent, the value of eliminating uncertainty will naturally decrease.

\begin{table*}[!ht]
\centering

\caption{Novel writing generalization experiment results. Dialogue efficiency and output quality are reported. All win-rate proportions are based on 80 pairwise judgments per model pair (20 prompts × 2 judges × 2 renderers).}

\label{tab:novel_generalization}
\setlength{\tabcolsep}{5pt} 
\begin{tabular}{lcc ccccccc}
\toprule
& Turns & Total IG & Coverage & \multicolumn{3}{c}{Human Judge(↑)} & \multicolumn{3}{c}{GPT-5 Judge(↑)} \\
\cmidrule(lr){5-7} \cmidrule(lr){8-10}
Method & (↓) & (↑) & (↑) & \footnotesize \textbf{Win} & \footnotesize \textbf{W/T(0.5)} & \footnotesize \textbf{W/T} & \footnotesize \textbf{Win} & \footnotesize \textbf{W/T(0.5)} & \footnotesize \textbf{W/T} \\
\midrule
Nous (OfG)   & 14.2 & \textbf{65.4} & \textbf{0.77} & \textbf{0.51} & \textbf{0.54} & \textbf{0.57} & \textbf{0.51} & \textbf{0.56} & \textbf{0.61} \\
Nous (SFT)   & 11.1 & 60.7 & 0.73                   & 0.49 & 0.51 & 0.53 & 0.44 & 0.51 & 0.57 \\
GPT-few     & \textbf{10.4 }& 57.8 & 0.68                   & 0.43 & 0.46 & 0.50 & 0.40 & 0.48 & 0.56 \\
GPT-zero    & 13.7 & 55.2 & 0.67          & 0.46 & 0.49 & 0.51 & 0.37 & 0.45 & 0.53 \\
\bottomrule
\end{tabular}
\vspace{-4mm}
\end{table*}

\section{Detailed Introduction to the VisPainter Framework and  Analysis of Experimental Results}
\label{sec:vispainter}

We incorporate the VisPainter framework to establish a quantitative evaluation pipeline, complementing the evaluations presented in Section~\ref{sec:exp2}. While subjective assessments by humans or AI focus on perceptual alignment, they lack granular quantification. VisPainter addresses this by providing an end-to-end process—from generation to execution—where output quality can be measured against specific, quantifiable indicators provided by VisBench. This integration serves as a critical supplement to our experiments, providing objective metrics to corroborate the effectiveness of our proposed method.

\paragraph{Introduction to the VisPainter Framework}
We adopt VisPainter as a baseline because it addresses a key limitation of diffusion-based text-to-image models: instead of producing rasterized bitmaps, it generates fully editable vector diagrams. This property is crucial for scientific illustration, where accuracy, semantic clarity, and iterative refinement are essential.

VisPainter is a multi-agent framework built on the Model Context Protocol (MCP), organized into three collaborative roles. The \textit{Manager} parses intent and coordinates tasks; the \textit{Designer} drafts and refines layouts; and the \textit{Toolbox} provides over thirty MCP-wrapped atomic drawing operations. In our experimental setup, GPT-4o serves as the Manager and Gemini-1.5-Pro as the Designer, while the Toolbox handles structured execution. These roles collaborate to translate natural language instructions into structured, editable diagrams through iterative refinement.

Furthermore, the evaluation module within VisPainter is VisBench, a benchmark designed for scientific schematics. It provides seven evaluation metrics across four dimensions: accuracy, recall, design error rate, blank space rate, readability, alignment, and design steps. In our evaluation, we focus strictly on output quality, excluding the ``design steps" metric. The VisBench dataset contains 360 entries, split evenly between T2I (Text-to-Image) and TI2I (Text-Image-to-Image) scenarios. Our 100 test sets are selected from the T2I subset. This integration transforms VisPainter from a generative system into a rigorous research platform, ensuring reproducible and fair benchmarking. To the best of our knowledge, VisPainter was developed concurrently with our work, and its open-source release is forthcoming.

\paragraph{In-depth Analysis of Metric Validity}
\label{para:metric_validity}
In the results of Experiment \ref{sec:exp2}, we observe a divergence in metric trends: while accuracy and recall improve significantly with our method, scores for design error rate and alignment show a slight decline. This phenomenon can be attributed to the intrinsic trade-off between information richness and execution complexity. Metrics such as \textit{Recall} and \textit{Accuracy} are positively correlated with information richness; as Nous captures more detailed constraints, the prompt becomes denser, naturally driving these scores higher. Conversely, metrics like \textit{Alignment} and \textit{Design Error Rate} are negatively correlated with task complexity. Since the capability of the backend designer (i.e., the plotting model) is fixed, increasing the number of components and structural constraints exponentially raises the execution difficulty. Untrained models often output simplistic diagrams with fewer elements, leaving little room for execution errors, which paradoxically results in higher ``stability" scores. Therefore, the slight drop in these specific metrics reflects the increased challenge of rendering high-fidelity diagrams rather than a failure of the inquiry agent. Ultimately, the significant gains in semantic accuracy outweigh these minor execution artifacts.

\section{Limitations and Future Work}
\paragraph{Limitations}

\textbf{The Attribute Independence Assumption:}
\label{sec:limitation}
In the main text, for computational tractability, we assume conditional independence between attributes. Although this is a reasonable and effective first-order approximation(as similar independence assumptions are also adopted in these works ~\citep{young2013pomdp,d_5,d_3,d_4,d_8}), many real-world tasks involve complex dependencies; for instance, a specific layout choice might constrain the types of available components. We acknowledge that ignoring these correlations may lead to an overestimation of entropy, causing the agent to adopt a more conservative strategy (\textit{e.g.}, asking redundant confirmatory questions), which may result in suboptimal query strategies of the model. However, this reduction to linear complexity is a necessary trade-off for proper operation ~\citep{d_2,d_6,d_7}, avoiding the exponential overhead of modeling full coupling. Furthermore, our hard-constraint update mechanism ensures robustness by forcing the posterior probability to collapse upon explicit user feedback, thereby restricting the cost of this assumption to minor efficiency losses rather than systemic intention misalignment. Our current model does not explicitly model these interactions, leaving this as a promising direction for future work.

\textbf{Scope of Application:} In this paper, we take the generation of scientific plots as our test scenario and have achieved good results. In addition, we have conducted preliminary experiments in the scenario of creative writing of novels, and the experimental results prove that our framework may have the potential for generalization. It is worth noting that our aim is to demonstrate the possibility of generalization of this method in specific scenarios, rather than claiming its absolute universality in all tasks. Furthermore, we have discussed in detail the applicability boundaries of our method under different interaction dimensions in the \ref{sec:Applicability Boundaries} section. Extending it to more diverse collaborative scenarios is an important direction for our future research.


\paragraph{Future Work}
The framework presented here has the potential to generalize to other structured domains, such as UI design, data visualization, or game creation. Beyond this broad applicability, two research directions are especially promising.

\textbf{Learning the Task Space:} Future agents could move beyond a fixed attribute set by inferring relevant attributes and their structure directly from interaction or large dialogue corpora. This would allow the framework to adapt dynamically to new tasks without manual specification.

\textbf{Toward Mixed-Initiative Dialogue:} Our current model is agent-led. A natural extension is to support mixed-initiative collaboration, where users proactively contribute information and the agent must decide whether to integrate it or pivot its strategy. This would yield more natural and adaptive interaction.

Together, these directions point toward making inquiry-driven collaboration more generalizable and human-like.

\section{Details of User Testing}
\label{sec:human}
We recruited 10 participants, all of whom hold a Bachelor’s degree or higher. Each participant was required to complete 100 tasks in two settings, resulting in a total of 2,000 interaction logs. During the experiments, participants were instructed to engage in natural language dialogue with the model based on a provided reference image. No constraints were placed on their linguistic expression; participants were encouraged to use either natural colloquialisms or professional terminology. Statistical results show that in the Zero-start setting, the average number of dialogue turns was $17.3 \pm 2.5$, with an average total Information Gain (IG) of $126.44 \pm 6.8$. In the Draft-start setting, the average number of turns was $7.6 \pm 1.8$, with an average total IG of $42.31 \pm 5.3$. A cross-comparison of the data from these 10 participants reveals remarkably minimal individual variance. This demonstrates that Nous is highly robust to variations in individual expression styles, effectively standardizing interaction efficiency across diverse users. These findings further confirm the statistical significance and experimental reliability of the conclusions presented in the main text.

\section{Implementation Details}
\label{sec:hayper}
\paragraph{Training Environment and Hyperparameters}
All models were trained using a full-parameter fine-tuning approach on a high-performance computing cluster equipped with 8x NVIDIA H200 (141GB) GPUs. We utilized bfloat16 mixed-precision training to optimize for speed and memory efficiency. The key hyperparameters used for training each of the models are detailed in Table \ref{tab:hyperparameters}. We selected these parameters based on preliminary experiments to ensure stable and effective training for each respective method.

\begin{table}[!ht]
\centering
\caption{Hyperparameters for SFT, DPO, OnG, and OfG.}
\label{tab:hyperparameters}
\begin{tabular}{lcccc}
\toprule
\textbf{Hyperparameter} & \textbf{SFT} & \textbf{DPO} & \textbf{OnG} & \textbf{OfG} \\
\midrule
\multicolumn{5}{l}{\textit{Model \& Data Configuration}} \\
Base Model & \multicolumn{4}{c}{Qwen3-8B} \\
Fine-tuning Method & \multicolumn{4}{c}{Full-parameter} \\
Training Precision & \multicolumn{4}{c}{bfloat16} \\
Max Sequence Length & \multicolumn{4}{c}{4096} \\
\midrule
\multicolumn{5}{l}{\textit{Optimization}} \\
Optimizer & \multicolumn{4}{c}{AdamW} \\
Learning Rate (lr) & 1e-6 & 1e-6 & 1e-6 & \textbf{1e-6} \\
LR Scheduler Type & \multicolumn{4}{c}{Cosine} \\
Warmup Steps & 50 & 50 & 50 & \textbf{50} \\
Epochs & 5 & 5 & 5 & \textbf{5} \\
Batch Size (per device) & 1 & 1 & 1 & \textbf{1} \\
Gradient Accum. Steps & 2 & 2 & 2 & \textbf{2} \\
Weight Decay & 0.01 & 0.01 & 0.01 & \textbf{0.01} \\
\midrule
\multicolumn{5}{l}{\textit{Regularization \& RL-specific}} \\
KL Coefficient ($\beta$) & N/A & 0.1 & 0.01 & \textbf{0.01} \\
PPO Clip Epsilon ($\epsilon$) & N/A & N/A & 0.2 & \textbf{0.2} \\
\bottomrule
\end{tabular}
\end{table}

\section{Dataset Details}
\label{sec:data_details}
Our dataset was constructed from a corpus of scientific papers sourced from arXiv and PubMed Central (PMC), covering a wide range of disciplines to ensure diversity. The primary arXiv categories included Computer Science (43.1\%), Physics (22.7\%), Quantitative Biology (14.8\%), Electrical Engineering (11.5\%), and others such as Economics and Statistics (7.9\%). All source materials were confirmed to be under open-access licenses (\textit{e.g.}, Creative Commons, arXiv.org non-exclusive license) that permit reuse for research. The initial pool of approximately 1 million figures was refined through a multi-stage pipeline: an initial filtering with CLIP to remove data plots, followed by a fine-grained selection of schematic diagrams using Qwen-2.5-VL-72B. A final manual verification by three domain experts ensured the relevance and quality of each diagram, resulting in a curated set of 1,100 figures. Of these, 1,000 were used for training and 100 were held out for testing. To ensure the reproducibility of our experiments involving proprietary models, all API calls for data generation and evaluation were made using model versions available after 4-14-2025.

\vspace{-1mm}
\section{prompts}
\vspace{-1mm}
All prompt templates, data construction, model training, and result evaluation codes are included in the attachments submitted along with the article. Here we provide the Socratic prompting templates used for our zero-shot and few-shot baselines. The core idea is to encourage the assistant to proactively ask clarifying questions before finalizing the diagram specification:

\vspace{-1mm}
\subsection{Zero-Shot Socratic Prompting}
\begin{lstlisting}
"You are an assistant that helps design scientific diagrams. 
Do not produce the diagram immediately. Instead, follow these steps:
1. Ask the user a clarifying question about the diagram (e.g., type, layout, number of components, connections, or style).
2. Continue asking such clarifying questions until enough information has been gathered to produce a complete diagram specification.
3. Only after clarification is complete, summarize the final diagram specification in a structured format (JSON).
Remember:
- Ask focused, concrete questions (one per turn).
- Avoid vague or open-ended questions.
- The final specification must be complete and self-contained."
\end{lstlisting}
\vspace{-1mm}
\subsection{Few-Shot Socratic Prompting}
\begin{lstlisting}
"You are an assistant that helps design scientific diagrams. 
Do not produce the diagram immediately. Instead, follow these steps:
1. Ask the user a clarifying question about the diagram (e.g., type, layout, number of components, connections, or style).
2. Continue asking such clarifying questions until enough information has been gathered to produce a complete diagram specification.
3. Only after clarification is complete, summarize the final diagram specification in a structured format (JSON).
Remember:
- Ask focused, concrete questions (one per turn).
- Avoid vague or open-ended questions.
- The final specification must be complete and self-contained.
Here are two example questions for your reference:Example 1: Example 2:"
\end{lstlisting}

\section{Data and Illustration Example}
\label{sec:data_example}
\paragraph{Data Example}

\textbf{Example of offline GRPO training data based on information entropy:}
\begin{lstlisting}
"  {
    "prompt": "User: I want to create a scientific diagram.\nAssistant: What specific element or component should serve as the target identifier (connection.to_id) for the connections ...",
    "responses": [
      "What is the primary scientific concept this diagram aims to illustrate?",
      "What is the primary scientific concept or model you intend to illustrate with these components and connections?",
      "What is the fundamental scientific concept or process that your diagram is intended to illustrate?",
      "What primary scientific concept or system does this diagram aim to illustrate?",
      "What is the primary scientific concept or system that this diagram is intended to illustrate?",
      "What overall visual structure do you envision for your diagram?",
      "Which overall diagram type do you envision such as a flowchart...?",
      "What specific element or component should ...?"
    ],
    "reward": [
      16.83,
      13.91,
      10.56,
      8.21,
      6.78,
      4.09,
      1.32,
      0.0
    ]
  },"
\end{lstlisting}

\textbf{Example of offline GRPO training data based on slot-counting:}
\begin{lstlisting}
"    {
    "prompt": "User: I want to create a scientific diagram.",
    "responses": [
      "What is the primary scientific concept this diagram aims to illustrate?",
      "What is the primary scientific concept or topic that this diagram is intended to illustrate?",
      "What is the primary scientific concept or topic that this diagram aims to illustrate?",
      "What are the main components of your scientific diagram, and what unique identifiers (component.id) will you assign to each?",
      "What specific components (component.id) do you envision in your diagram, and what identifiers or labels should be assigned to each to clarify their roles?",
      "Which key components (nodes) do you envision for your diagram, and how would you uniquely identify each (i.e., what are their respective component IDs)?",
      "Can you identify the distinct components for your diagram by assigning specific IDs or names, and briefly describe the role of each?",
      "What are the main components (component.id) you envision including in your scientific diagram, and what specific role does each play in illustrating the concept?"
    ],
    "reward": [
      4.0,
      2.0,
      2.0,
      1.0,
      1.0,
      1.0,
      1.0,
      1.0
    ]
  },"
\end{lstlisting}

\begin{figure*}[ht]
    \centering
    \includegraphics[width=0.97\linewidth]{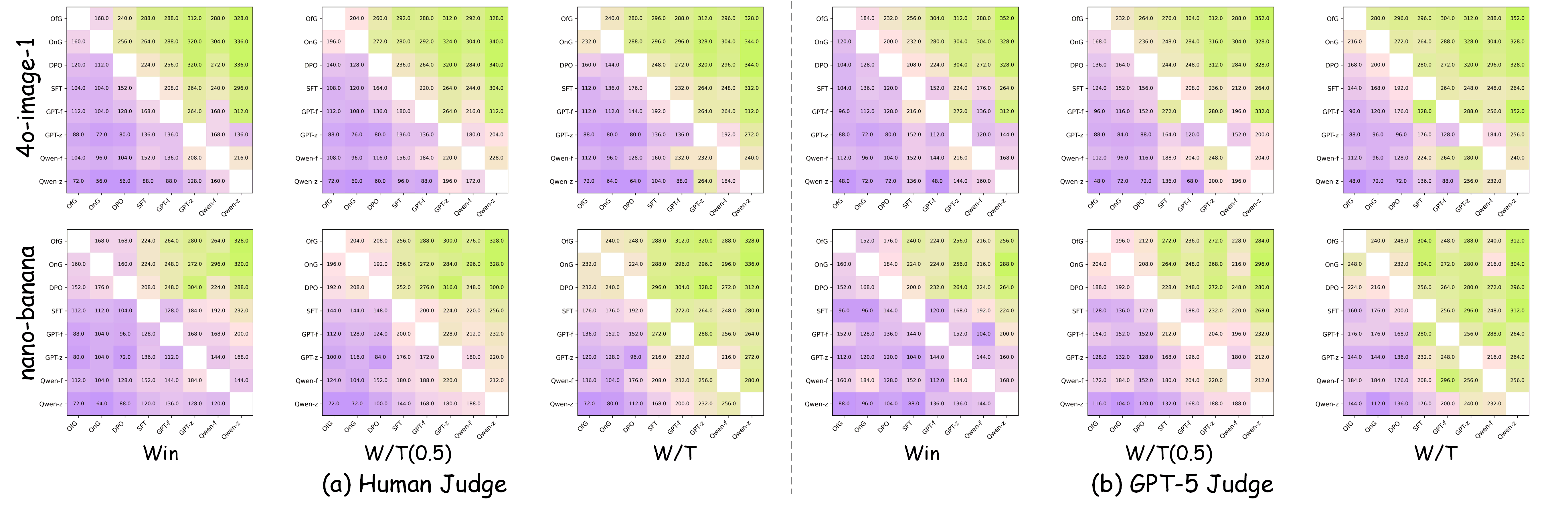}
    \caption{Model scores under different tie-handling protocols. (a) Results of human evaluation; (b) Results of GPT-5 model evaluation.}
    \label{fig:heatmap}
\end{figure*}

\paragraph{Heatmap of evaluation results}
\label{sec:heatmap_figure}
The heatmap(Figure ~\ref{fig:heatmap}) of winning rates of different experimental groups in manual evaluation and GPT-5 evaluation in the main text section \ref{sec:exp2}.

\begin{figure}[ht]
    \centering
    \includegraphics[width=\linewidth]{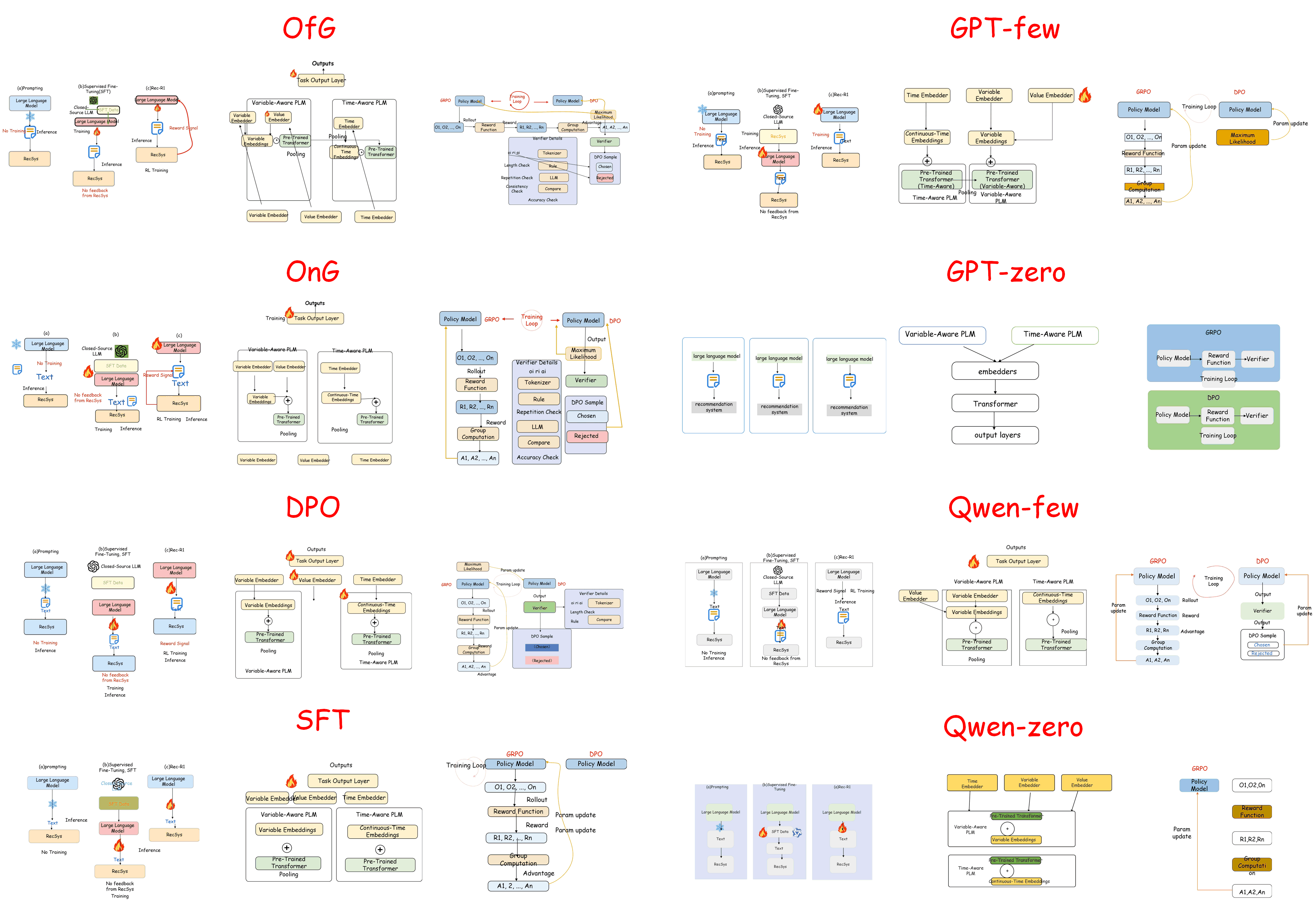}
    \caption{This section presents drawing examples generated using the VisPainter framework}
    \label{fig:vispng}
    \vspace{-2mm}
\end{figure}

\paragraph{Objective metrics}
\label{sec:vispainter_figure}
This section presents drawing examples generated using the VisPainter framework, as shown in Figure \ref{fig:vispng}.

\begin{figure}[!ht]
    \centering
    \includegraphics[width=0.85\linewidth]{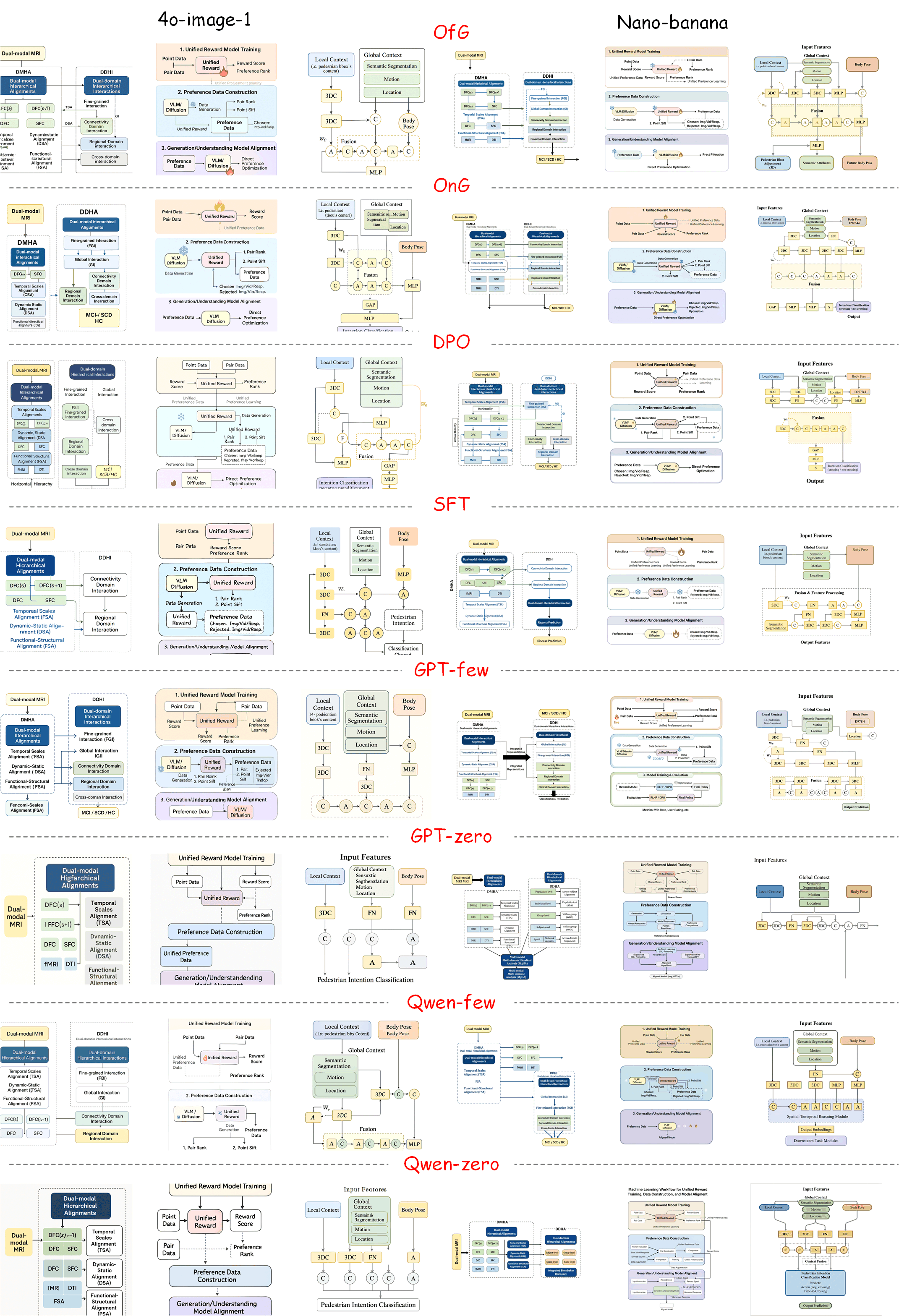}
    \caption{Partial Examples of Model-Generated Images}
    \label{fig:modelpng}
    \vspace{-2mm}
\end{figure}

\paragraph{Subjective comparison}
\label{sec:model_figure}
This section presents drawing examples generated by two models (4o-image-1 and nano-banana), as shown in Figure \ref{fig:modelpng}.


\end{document}